\documentclass{article}

\usepackage[final]{neurips_2021}

\usepackage{amsmath,amsfonts,bm}

\def\eqref#1{equation~\ref{#1}}
\def\1{\bm{1}}

\def\va{{\bm{a}}}

\def\vc{{\bm{c}}}

\def\vi{{\bm{i}}}
\def\vj{{\bm{j}}}
\def\vk{{\bm{k}}}

\def\vo{{\bm{o}}}
\def\vp{{\bm{p}}}
\def\vq{{\bm{q}}}
\def\vr{{\bm{r}}}

\def\vt{{\bm{t}}}

\def\vx{{\bm{x}}}

\DeclareMathAlphabet{\mathsfit}{\encodingdefault}{\sfdefault}{m}{sl}
\SetMathAlphabet{\mathsfit}{bold}{\encodingdefault}{\sfdefault}{bx}{n}

\DeclareMathOperator*{\argmin}{arg\,min}

\usepackage[utf8]{inputenc} % allow utf-8 input
\usepackage[T1]{fontenc}    % use 8-bit T1 fonts
\usepackage{hyperref}       % hyperlinks
\usepackage{url}            % simple URL typesetting
\usepackage{booktabs}       % professional-quality tables
\usepackage{amsfonts}       % blackboard math symbols
\usepackage{nicefrac}       % compact symbols for 1/2, etc.
\usepackage{microtype}      % microtypography
\usepackage{xcolor} % colors
\usepackage{float}

\usepackage{microtype}
\usepackage{subfigure}
\usepackage{booktabs} % for professional tables
\usepackage{algorithm2e}
\usepackage{wrapfig}
\usepackage[export]{adjustbox}
\usepackage{algorithm}
\usepackage{algorithmic}
\usepackage{bm}
\newcommand{\modelname}{{\textsc{NPS}}}

\usepackage{multirow}

\usepackage{amsmath}
\usepackage{todonotes}

\usepackage{url}
\usepackage[utf8]{inputenc} % allow utf-8 input
\usepackage[T1]{fontenc}    % use 8-bit T1 fonts
\usepackage{hyperref}       % hyperlinks
\usepackage{cleveref}
\usepackage{algorithm}

\usepackage{url}            % simple URL typesetting
\usepackage{booktabs}       % professional-quality tables
\usepackage{amsfonts}       % blackboard math symbols
\usepackage{nicefrac}       % compact symbols for 1/2, etc.
\usepackage{microtype}      % microtypography

\newcommand{\bull}{{\tiny $\bullet~~$ }}
\usepackage{caption}

\newcommand{\vM}{{\bm{M}}}
\newcommand{\vN}{{\bm{N}}}
\newcommand{\vT}{{\bm{T}}}

\newcommand{\vX}{{\bm{X}}}
\newcommand{\vW}{{\bm{W}}}

\newcommand{\vV}{{\bm{V}}}

\newcommand{\vR}{\bm{R}}
\newcommand{\vP}{\bm{P}}

\usepackage{pifont}% http://ctan.org/pkg/pifont

\newcommand{\highlight}[1]{\colorbox{blue!10}{#1}}
\definecolor{mygray}{gray}{0.4}

\newcommand{\g}[2]{#1\textsubscript{\textcolor{mygray}{$\pm$#2}}}
\author{
Aniket Didolkar\textsuperscript{*, 1}, 
Anirudh  Goyal \textsuperscript{*, 1},
Nan Rosemary Ke \textsuperscript{2},
Charles Blundell \textsuperscript{2}, \\
\textbf{Philippe Beaudoin \textsuperscript{3}
Nicolas Heess \textsuperscript{2},
Michael Mozer \textsuperscript{4},
Yoshua Bengio \textsuperscript{1}}}

\DeclareUnicodeCharacter{2212}{-}

\title{Neural Production Systems}

\begin{document}

\let\footnote\relax\footnotetext{ \textsuperscript{*} Equal Contribution,  \textsuperscript{1} Mila, University of Montreal, \textsuperscript{2} Google Deepmind, \textsuperscript{3} Waverly, \textsuperscript{4} Google Research, Brain Team
Corresponding authors: \texttt{anirudhgoyal9119@gmail.com}, \texttt{adidolkar123@gmail.com} 
}

%\textsuperscript{*} Equal Contribution, \textsuperscript{1} Mila, University of Montreal, \textsuperscript{2} Google Deepmind, \textsuperscript{3} Waverly, \textsuperscript{4} Google Brain, 
%Corresponding authors: \texttt{anirudhgoyal9119@gmail.com}}

%\let\thefootnote\relax\footnotetext{\textsuperscript{*} Equal Contribution, \textsuperscript{**} Equal Advising \textsuperscript{1} Mila, University of Montreal, \textsuperscript{2} Google Deepmind, \textsuperscript{3} Waverly, \textsuperscript{4} Google Research, Brain Team. 
%Corresponding authors: \texttt{anirudhgoyal9119@gmail.com}, \texttt{adidolkar123@gmail.com} 
%}

\maketitle

\begin{abstract}
  Visual environments are structured, consisting of distinct  objects or \emph{entities}. These entities have properties---visible or latent---that determine the manner in which they interact with one another. To partition images into entities, deep-learning researchers have proposed structural inductive biases such as slot-based architectures. To model interactions among entities, equivariant graph neural nets (GNNs) are used, but these are not particularly well suited to the task for two reasons. First, GNNs do not predispose interactions to be sparse, as relationships among independent entities are likely to be.  Second, GNNs do not factorize knowledge about  interactions in an entity-conditional manner. As an alternative, we take inspiration from cognitive science and resurrect a classic approach, \emph{production systems}, which consist of a set of rule templates that are applied by binding placeholder  variables in the rules to specific entities. Rules are scored on their match to entities, and the best fitting rules are applied to update entity properties. In a series of experiments, we demonstrate that this architecture achieves a flexible, dynamic flow of control and serves to factorize entity-specific and rule-based information. This disentangling of knowledge achieves robust future-state prediction in rich visual environments, outperforming state-of-the-art methods using GNNs, and allows for the extrapolation from simple (few object) environments to more complex environments.

\end{abstract}

\section{Introduction}

Despite never having taken a physics course, every child beyond a young age appreciates that pushing a plate off the dining table will cause the plate to break. The laws of physics accurately characterize the dynamics of our natural world, and although explicit knowledge of these laws is not necessary to reason, we can reason explicitly about objects interacting through these laws. Humans can verbalize knowledge in propositional expressions such as 
``If a plate drops from table height, it will break,''  and
``If a video-game opponent approaches from behind and they are carrying a weapon, they are likely to attack you.''
Expressing propositional knowledge is 
not a strength of current  deep learning methods for several reasons. First, propositions are discrete and independent from one another.  Second, propositions must be  quantified in the manner of first-order logic; for example, the video-game proposition applies to any $X$ for which $X$ is an opponent and has a weapon. Incorporating the ability to express and reason about propositions should improve generalization  in deep learning methods because this knowledge is modular--- propositions can be formulated independently of each other---
and can therefore be acquired incrementally.
Propositions can also be composed with each other and applied consistently to all entities that match, yielding a powerful form
of \emph{systematic generalization}.

\iffalse
\begin{figure}[h]
    \begin{center}
    \vspace*{-3mm}
\begin{minipage}{5cm}
\begin{minted}[linenos, frame=lines,
framesep=4mm,baselinestretch=1.2]{cpp}
int add(int a, int b){
   int c = a + b;
   return c;
}
add(4,2); // 6
add(1024, 2048); // 3072
\end{minted}
\end{minipage}
\vspace*{-3mm}
\end{center}
    \caption{Example of typed argument function in C++. Here, function \textit{add} can be applied to any two entities as long as their type matches the expected types (in this case, integer). Following this analogy, we aim to decompose knowledge into a neural form of typed production rules and entities, i.e., with distributed representations for types and entities and MLPs as functions.}
    \label{fig:code_example}
\end{figure}
\fi
%\Ani{
%Connectionist networks are good at performing pattern matching, especially when there is no perfect match and the aim is to find the best partial match. 

%This falsifies any strong claim that connectionist systems using distributed representations could not possibly implement symbol processing. 

%taatgen2003production, anderson1996architecture, anderson2004integrated,anderson2014rules
The classical AI literature from the 1980s can offer deep learning researchers a valuable perspective. In this era, reasoning, planning, and prediction were handled by architectures that performed propositional inference on symbolic knowledge representations. A simple example of such an architecture is the  \emph{production system} \citep{laird1986chunking,  anderson1987skill}, which expresses knowledge by \emph{condition-action rules}. The rules operate on a \emph{working memory}: rule conditions are matched to entities in working memory inspired by cognitive science, and such a match can trigger computational actions that update working memory or  external actions that operate on the outside world. 

Production systems were typically used to model high-level cognition, e.g., mathematical problem solving or procedure following;  perception was not the focus of these models. It was assumed that the results of perception were placed into working memory in a symbolic form that could be operated on with the rules. In this article, we revisit production systems but from a deep learning perspective which naturally integrates perceptual processing and subsequent inference for visual reasoning problems. We describe an end-to-end deep learning model that constructs object-centric representations of entities in videos, and then operates on these entities with differentiable---and thus learnable---production rules. The essence of these rules, carried over from traditional symbolic system, is that they operate on variables that are \emph{bound}, or linked, to  the entities in the world. In the deep learning implementation, each production rule is represented by a distinct MLP with  query-key attention mechanisms to specify the rule-entity binding and to determine when the rule should be triggered for a given entity. We are not the first to propose a neural instantiation of a production system architecture. \citet{touretzky1988distributed} gave a proof of principle that neural net hardware could be hardwired to implement a production system for symbolic reasoning; our work fundamentally differs from theirs in that (1) we focus on perceptual inference problems and (2) we use the architecture as an inductive bias for learning.

%The inner workings of production rules themselves do not have to be reducible to a symbolic explanation, but the rules could be explicated: although humans cannot always explain all the details of how an action is carried out, they can name it along with the characteristics of variables on which it acts or has an effect on. Hence in the setup explored here, a production rule is represented by a small MLP, with query-key attention mechanisms to specify the characteristics of its arguments and when the rule should be triggered.

%%%%ANI: Add it again.

%\vspace{-3mm}
\subsection{Variables and entities}
%\vspace{-3mm}

%In theories of conscious processing like the Global Workspace theory~\citep{baars1997theatre,Dehaene-et-al-2017}, entities extracted from the sensory signals compete for access to consciousness and an attention mechanism selects among these entities detected at an non-conscious level those which enter into conscious reasoning steps which bind these entities with generic pieces of knowledge about them, which we associate here with neural production rules.

\begin{figure}
    \centering
    \includegraphics[width = 14cm]{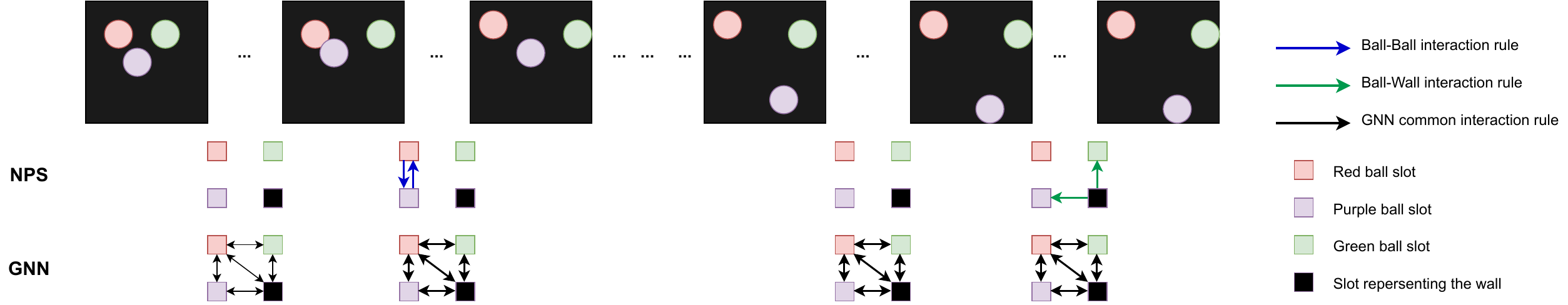}
    \caption{In this figure we show a visual comparison between NPS and dense architectures like GNNs. In NPS, a rule is only applied when an interaction takes place and it is applied only to the slots affected by the interaction. NPS also uses different rules for different kinds of interactions, while in GNN a common rule is applied to all slots irrespective of whether an interaction takes place or not (because of parameter sharing). Note the dynamic nature of the interaction graph in NPS, while in GNN, the graph is static. }
    \label{fig:nps_diag}
    \vspace{-2\baselineskip}
\end{figure}

\iffalse
\begin{wrapfigure}{r}{0.50\textwidth}
\vspace{-1.0\baselineskip}

    \includegraphics[width=0.48\textwidth]{figures/SS/ss_rule_vis.pdf}
  \caption{Here we show the rule selection statistics from the proposed model for all entities in the shapes stack dataset across all examples. Each example contains 3 entities as shown above. Each cell in the table shows the probability with which the given rule is triggered for the corresponding entity. We can see that the bottom-most entity triggers rule 2 most of the time while the other 2 entities trigger rule 1 most often. This is quite intuitive as, for most examples, the bottom-most entity remains static and does not move at all while the upper entities fall. Therefore, rule 2 captures information which is relevant to static entities, while rule 1 captures physical rules relevant to the interactions and motion of the upper entities.} \label{fig:ss_demo}
      \vspace{-1\baselineskip}
\end{wrapfigure}
\fi

What makes a rule general-purpose is that it incorporates placeholder \emph{variables} that can be bound to arbitrary \emph{values} or---the term we prefer in this 
article---\emph{entities}. This notion of binding is familiar in functional programming languages, where these variables
are called arguments. Analogously, the use of variables in the production rules we describe enable a model to reason about any set of entities that satisfy the selection criteria of the rule.

%% ADD This back in.
Consider a simple function in C like \verb+int add(int a, int b)+. This function binds its two integer operands to variables $a$ and $b$. The function does not apply if the operands are, say, character strings. The use of variables enables a  programmer to reuse the same function to add any two integer values

%%Without a factorization of image data into entities, it would not be possible to bind variables (the arguments of neural production rules) to entities (observed in the visual fields).  

In order for rules to operate on entities, these entities must be represented explicitly. That is, the visual world needs to be  parsed in a task-relevant manner, e.g., distinguishing the sprites in a video game or the vehicles and pedestrians approaching an  autonomous vehicle. Only in the past few years have deep learning vision researchers developed methods for object-centric representation \citep{le2011learning, eslami2016attend, greff2016tagger, raposo2017discovering, van2018relational, kosiorek2018sequential, engelcke2019genesis, burgess2019monet, greff2019multi, locatello2020object, ahmed2020causalworld, goyal2019recurrent, zablotskaia2020unsupervised, rahaman2020s2rms, du2020unsupervised, ding2020object, goyal2020object, ke2021systematic}. These methods differ in details but share the notion of a  fixed number of \emph{slots} (see Figure \ref{fig:nps_diag} for example), also known as \emph{object files}, each encapsulating information about a single object. Importantly, the slots are interchangeable, meaning that it doesn't matter if a scene with an apple and an orange encodes the apple in slot 1 and orange in slot 2 or vice-versa. 

A model of visual reasoning must not only be able to represent entities but must also express knowledge about 
entity dynamics and interactions. To ensure \emph{systematic} predictions, a model must be capable of applying knowledge
to an entity regardless of the slot it is in and must be capable of applying the same knowledge to multiple instances of an entity.
Several distinct approaches exist in the literature. The predominant approach uses graph neural networks to model slot-to-slot 
interactions \citep{scarselli2008graph, bronstein2017geometric, watters2017visual,  van2018relational, kipf2018neural, battaglia2018relational, tacchetti2018relational}. To ensure systematicity, the GNN must share parameters among the edges.
In a recent article, \citet{goyal2020object} developed a more general framework in which parameters are shared but
slots can dynamically select which parameters to use in a state-dependent manner.  Each set of parameters is referred to as a \emph{schema}, and slots use a query-key attention mechanism to select which schema to apply at each time step. Multiple
slots can select the same schema. 
In both GNNs and SCOFF, modeling dynamics involves each slot interacting with each other slot. In the work we describe in this
article, we replace the direct slot-to-slot interactions with rules, which mediate sparse interactions among slots (See arrows in Figure \ref{fig:nps_diag}).

Thus our main contribution is that we introduce NPS, which offers a way to model dynamic and sparse interactions among the variables in a graph and also allows dynamic sharing of multiple sets of parameters among these interactions. Most architectures used for modelling interactions in the current literature use statically instantiated graph which model all possible interactions for a given variable at each step i.e. dense interactions. Also such dense architectures share a single set of parameters across all interactions which maybe quite restrictive in terms of representational capacity.  A visual comparison between these two kinds of architectures is shown in Figure \ref{fig:nps_diag}. Through our experiments we show the advantage of modeling interactions in the proposed manner using NPS in visually rich physical environments. We also show that our method results in an intuitive factorization of rules and entities. 
%Through our experiments, we show that factorization of knowledge into rules provides a strong inductive bias for learning interaction dynamics among entities in rich physical environments. Representing entity dynamics using \modelname\ leads to impressive performance gains over commonly used models such as GNNs in a wide variety of physical environments. We also find that the distinct rules learned by the proposed model are quite intuitive and interpretable as shown in Figure \ref{fig:ss_demo}. The figure shows rule selection statistics of the proposed Neural Production System model for the shapes stack dataset \citep{shapestack} when using 2 rules. The shapes stack dataset consists of 3 entities that are stacked into a tower and fall under the influence of gravity. In the next section, we describe the proposed model that learns neural representations of entities and rules.

\vspace{-3mm}
%\iffalse
\section{Production System}
%\vspace{-2mm}

\label{desiderata}

% %Conditions may reflect an aspect of the external world (e.g., it is raining outside) or an internal mental state (e.g., to retrieve a particular fact). Likewise, actions may transform a feature in the real world (e.g., use an umbrella if its raining) or an internal mental state (e.g., add a fact to memory).
%Information about the state of the entities is matched to all productions and if the condition on the left of the arrow is true, then the action on the right can be performed. 

Formally, our notion of a production system consists of a set of entities and a set of rules, along with a mechanism for selecting rules to apply on subsets of the entities. Implicit in a rule is a specification of the properties of relevant entities, e.g.,
a rule might apply to one type of sprite in a video game but not another.  The control flow of a production system dynamically selects rules as well as bindings between rules and entities, allowing different rules to be chosen and different entities to
be manipulated at each point in time.

The neural production system we describe shares essential properties with traditional production system, particularly with
regard to the compositionality and generality of the knowledge they embody. \citet{lovett2005thinking} describe four desirable
properties commonly attributed to symbolic systems that apply to our work as well.

\textit{Production rules are modular}. Each production rule represents a unit of knowledge and are \textit{atomic} such that any production rule can be intervened (added, modified or deleted) independently of other production rules in the system.

\textit{Production rules are abstract.} Production rules allow for generalization because their conditions may be represented as high-level abstract knowledge that match to a wide range of patterns. These conditions specify the attributes of relationship(s) between entities without specifying the entities themselves. The ability to represent  abstract knowledge allows for the transfer of learning across different environments as long as they fit within the conditions of the given production rule.

%For ex. a static rule is a thing like 'if Bob is hungry then he looks for food'. Instead, a more general production rule is a thing like, 'for all $X$, if $X$ is a human and $X$ is hungry, then $X$ looks for food' (with some probability). $X$ can be bound to specific instances (or to other variables which may involve more constraints on the acceptable set). 
 % In classical AI, we have a thing called unification to keep track of how variables can be 'bound' to instances or to expressions (which mean some more constraints on which instances can be matched), when exploring whether some rule can be applied to some facts (in our database of facts).

%That the joint distribution over the entities is represented by a sparse factor graph.

\textit{Production rules are sparse. } In order that production rules have broad applicability, they involve only a subset of entities. This assumption imposes a strong prior that dependencies among entities are sparse. In the context of visual reasoning, we conjecture that this prior is superior to what has often been assumed in the past, particularly in the disentanglement literature---independence among entities  \cite{higgins2016beta, chen2018isolating}.
%lead to a structured prior where the application of a particular rule relates only a sparse subset of the entities. When learning from visual data, this prior encourages sparse dependencies over the entities. We conjecture that this prior is superior to what has been supposed in the past---independence among entities, as is often
%assumed in approaches to disentanglement 

%Instead of imposing a very strong prior of complete independence at the highest level of representation, we can have this slightly weaker but very structured prior, that the joint is represented by a sparse factor graph.

\textit{Production rules represent causal knowledge and are thus asymmetric.} 
Each rule can be decomposed into a \{condition, action\} pair, where the action reflects a state change that is a  causal consequence of the conditions being met.
%Each production rule is a unidirectional contingency for an action. This means that the production rule “When I want to type the letter ‘j’, then I punch my right index finger” is different from “When I punch my right index finger, then I type the letter ‘j’”. %Moreover, asymmetry and modularity imply that, if these two production rules were in the same system, adding, deleting or refining the former would not directly change the latter. That is, practicing typing would exercise the first production rule, strengthening the index finger response when “j” is the desired letter, but it would not strengthen one’s knowledge that “j” appears when touch-typing with that finger.  %\Ani{Relate to cause and effect. }

%\textit{Production rules are not directly verbalizable.} This feature is based on the notion that each production rule represents knowledge about a contingency of an action that is not directly accessible to verbalization.  It is important to note that, while this feature implies that the knowledge represented in the production rule form cannot be accessed directly, it does not imply that one cannot use other techniques to talk about performance knowledge.  Moreover, knowledge about how to perform a task may be represented in multiple forms, some verbalizable and some not.  %\Ani{This desiderata is not directly relevant for us in this paper..}

These four properties are sufficient conditions for knowledge to be expressed in production rule form.
These properties specify \emph{how} knowledge is represented, but not \emph{what} knowledge is represented.
The latter is inferred by learning mechanisms under the inductive bias provided by the form of production rules.
%\fi
%slot == placeholder.
%entity == contents

\vspace{-4mm}
\section{Neural Production System: Slots and Sparse Rules}
\vspace{-4mm}

\begin{wrapfigure}{r}{0.5\textwidth}
   %\centering
   \vspace{-1.5\baselineskip}
   \centering
   \includegraphics[width=.5\linewidth]{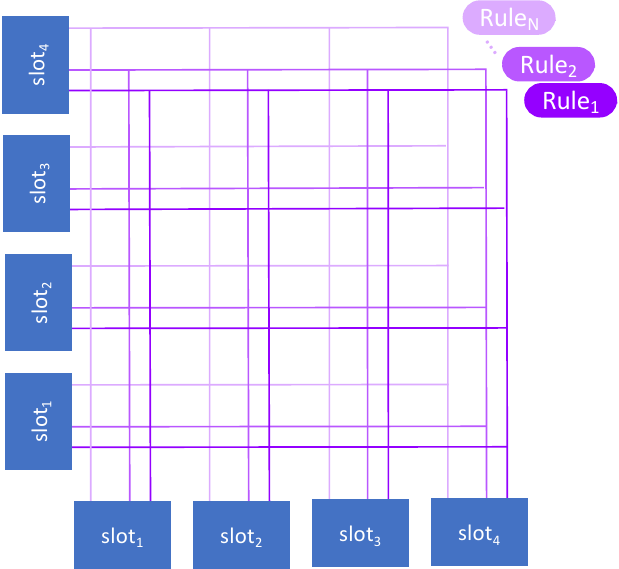}
    \caption{\textbf{Rule and slot combinatorics.} Condition-action rules specify how entities interact. Slots maintain the
    time-varying state of an entity. Every rule is matched to every pair of slots. Through key-value attention, a goodness
    of match is determined, and a rule is selected along with its binding to slots.}
   \label{fig:schematic_diagram}
    \vspace{-2.5\baselineskip}
\end{wrapfigure}

The Neural Production System (NPS), illustrated in Figure \ref{fig:schematic_diagram}, provides an architectural backbone that supports the detection and inference of entity (object) representations in an input sequence, and the underlying rules which govern the interactions between these entities in time and space. The input sequence indexed by time step $t$, $\{ \vx^1, \ldots, \vx^t, \ldots,  \vx^T\}$, for instance the frames in a video, are processed by a neural encoder \citep{burgess2019monet, greff2019multi, goyal2019recurrent, goyal2020object} applied to each $\vx^{t}$, to obtain a set of $M$ entity representations  $\{ \vV_1^{t}, \ldots, \ldots, \vV_M^{t}\}$, one for each of the $M$ slots. These representations describe an entity and are updated based on
both the previous state, $\vV^{t-1}$ and the current input, $\vx^{t}$.

%These production rules are active processing components that update the internal state of the different entities represented by different slots at each time-step.

% sampled initially from a normal distribution
\modelname\ consists of $N$ separately encoded rules,
$\{\vR_1, \vR_2,.., \vR_N\}$.  
Each rule consists of two components, $\vR_i = (\Vec{\vR_i}, MLP_{i})$, 
where $\Vec{\vR_i}$ is a learned rule embedding vector, which can be thought of as a template defining the condition for when
a rule applies; and  $MLP_{i}$, which determines the action taken by a rule.
Both $\Vec{\vR_i}$ and the parameters of $MLP_{i}$ are learned along with the other parameters of the model using back-propagation on an objective optimized end-to-end.

In the general form of the model, each slot selects a rule that will be applied to it to change its state. This can potentially be performed several times, with possibly different rules applied at each step. Rule selection is done using an attention mechanism described in detail below. Each rule specifies conditions and actions on a pair of slots. Therefore, while modifying the state of a slot using a rule, it can take the state of another slot into account. The slot which is being modified is called the primary slot and other is called the contextual slot. The contextual slot is also selected using an attention mechanism described in detail below. 

%To reduce the complexity of the search, we assume that the two slots are asymmetric: one slot
%is \emph{primary} in that it is used both to match the rule condition and it is acted on by the rule; the other slot
%is \emph{contextual} in that it determines how the primary slot is acted upon. With this set up, we perform selection
%in two operations, first considering all combinations of \{rule, primary slot\}, making a selection, and then conditioned on
%the selection, choosing a contextual slot. The resulting search is reduced to $NM + M$ combinations.

%\textbf{Computational Steps.}
\vspace{-3mm}
\subsection{Computational Steps in \modelname}
\vspace{-3mm}
\label{section:computational_steps}
In this section, we give a detailed description of the rule selection and application procedure for the slots. First, we will formalize the definitions of a few terms that we will use to explain our method. We use the term \textbf{primary slot} to refer to slot $\vV_p$ whose state gets modified by a rule $\vR_r$. We use the term \textbf{contextual slot} to refer to the slot $\vV_c$ that the rule $\vR_r$ takes into account while modifying the state of the primary slot $\vV_p$.

\textbf{Notation.} We consider a set of $\vN$ rules $\{\vR_1, \vR_2, \ldots, \vR_N\}$ and a set of $\vT$ input frames $\{\vx^1, \vx^2, \ldots, \vx^T\}$. Each frame $\vx^t$ is encoded into a set of $\vM$ slots $\{\vV^t_1, \vV^t_2, \ldots, \vV^t_M\}$. In the following discussion, we omit the index over $\vt$ for simplicity.
%Algorithm~\ref{alg:neural_production_systems} formalizes \modelname. The algorithm involves a loop over time steps $t$. Within a time step, exactly $K$ rule applications occur. (Flexibility in the number of
%rule applications per step can be easily incorporated into the model.) Because each rule application can affect slot contents, we  need to add notation to index the entity representation for slot $j$, $\vV_j$, by both the time step $t$ and the number of 
%rules that have been applied during the time step, denoted $h$, i.e., $\vV_j^{t,h}$. When we omit $h$, we imply that
%we are referencing the state following all rule applications, $h=K$.

{\bfseries \itshape Step 1.} is external to \modelname\ and involves parsing an input image, $\vx^{t}$, into slot-based entities 
conditioned on the previous state of the slot-based entities.
Any of the methods proposed in the literature to obtain a slot-wise representation of
entities can be used \citep{burgess2019monet, greff2019multi, goyal2019recurrent, goyal2020object}. The next three steps constitute the rule selection and application procedure.

{\bfseries \itshape Step 2.} For each primary slot $\vV_p$, we attend to a rule $\vR_r$ to be applied. Here, the queries come from the primary slot: $\vq_p = \vV_pW^q$, and the keys come from the rules: $\vk_i = \Vec{R_i}W^k \quad \forall \vi \in \{1, \ldots, \vN\}$. The rule is selected using a straight-through Gumbel softmax \citep{jang2016categorical} to achieve a learnable hard decision: $\vr = \mathrm{argmax}_i(\vq_p\vk_i + \gamma), \text{ where } \gamma \sim \mathrm{Gumbel} (0,1)$. This competition is a noisy version of rule matching and prioritization in traditional production systems. %selects the \{rule, primary slot\} pair, determined on-the-fly by an attention-based soft competition. This competition is analogous to rule matching and prioritization in traditional production systems.
%The current state $\vV_j^{t,h}$ of a slot constructs a key that is matched to a query obtained from a rule embedding $\Vec{\vR_i}$; based on the goodness of the match, a rule and primary slot, indexed by ${r}$ and ${p}$, respectively, are selected. Straight-through Gumbel softmax \citep{jang2016categorical} is used for achieving a learnable hard decision. 

{\bfseries \itshape Step 3.} For a given primary slot $\vV_p$ and selected rule $\vR_r$, a contextual slot $\vV_c$ is selected using another attention mechanism. In this case the query comes from the primary slot: $\vq_p = \vV_pW^q$, and the keys from all the slots: $\vk_j = \vV_jW^q \quad \forall \vj \in \{1, \ldots, \vM\}$. The selection takes place using a straight-through Gumbel softmax similar to step 2: $\vc = \mathrm{argmax}_j(\vq_p\vk_j + \gamma), \text{ where } \gamma \sim \mathrm{Gumbel} (0,1)$. Note that each rule application is sparse since it takes into account only 1 contextual slot for modifying a primary slot, while other methods like GNNs take into account all slots for modifying a primary slot. %selects a contextual slot given the chosen primary slot and rule. The selected contextual slot is indexed by ${c}$. Again, straight-through Gumbel softmax is used for this competition among slots. 

{\bfseries \itshape Step 4. Rule Application:} the selected rule $\vR_r$ is applied to the primary slot $\vV_p$ based on the rule and the current contents of the primary and contextual slots. The rule-specific  $MLP_{{r}}$, takes as input the concatenated representation of the state of the primary and contextual slots, $\vV_{{p}}$ and $\vV_{{c}}$, and produces an output, which is then used to change the state of the primary slot $\vV_{{p}}$ by residual addition.

%There is one such \emph{rule-application} per stage, and it changes the state of the primary slot. We select only one rule per rule application to ensure the modularity condition mentioned in section \ref{desiderata}. Selecting one rule per rule application inhibits the composition of rules. To counteract this, we allow this entire process to be performed across multiple times (loop over $p$ in Algorithm~\ref{alg:neural_production_systems}), for each time step of the input sequence. For example, if we choose to perform the above process $K$ times per time-step $t$, then the procedure consisting of \emph{slot-rule pair selection} and \emph{rule-application} is done $K$ times \emph{sequentially}. In the context of \modelname, $K$ is the number of rule application stages, $M$ is the number of slots, $N$ is the number of rules and acts as a hyper-parameter.

\vspace{-3mm}
\subsection{Rule Application: Sequential vs Parallel Rule Application}
\vspace{-3mm}
In the previous section, we have described how each rule application only considers another contextual slot for the given primary slot i.e., \textbf{contextual sparsity}. We can also consider \textbf{application sparsity}, wherein we use the rules to update the states of only a subset of the slots. In this scenario, only the selected slots would be \textit{primary slots}. This setting will be helpful when there is an entity in an environment that is stationary, or it is following its own default dynamics unaffected by other entities. Therefore, it does not need to consider other entities to update its state. We explore two scenarios for enabling application sparsity.

\begin{wrapfigure}{r}{0.50\textwidth}
\vspace{-1.1\baselineskip}

    \includegraphics[trim = 4cm 10cm 0 0, clip, width=0.48\textwidth]{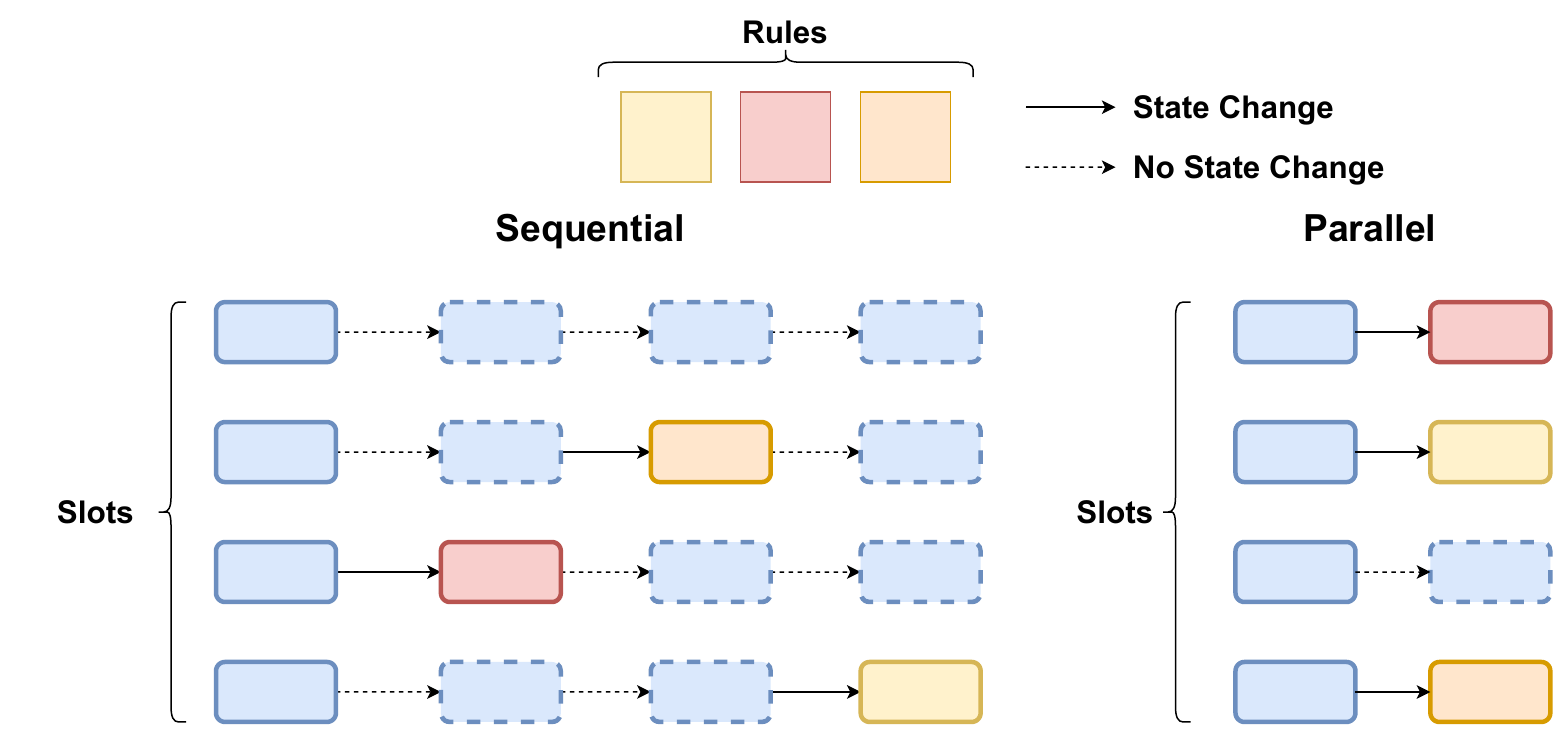}
  \caption{This figure demonstrates the sequential and parallel rule application.} \label{fig:seq_par}
      \vspace{-1.3\baselineskip}
\end{wrapfigure}

%Fixed.
% ha, nice!

\textbf{Parallel Rule Application.}  Each of the $M$ slots selects a rule to potentially change its state. To enable sparse changes,  we provide an extra \textbf{Null Rule} in addition to the available $N$ rules. If a slot picks the null rule in step 2 of the above procedure, we do not update its state. 

\textbf{Sequential Rule Application.} In this setting, only one slot gets updated in each rule application step. Therefore, only one slot is selected as the primary slot. This can be facilitated by modifying step 2 above to select one \{primary slot, rule\} pair among $\vN\vM$ \{rule, slot\} pairs. The queries come from each slot: $\vq_j = \vV_jW^q \quad \forall \vj \in \{1, \ldots, \vM\}$, the keys come from the rules: $\vk_i = R_iW^k \quad \forall \vi \in \{1, \ldots, \vN\}$. The straight-through Gumbel softmax selects one (primary slot, rule) pair: $\vp,\vr = \mathrm{argmax}_{i, j}(\vq_p\vk_i + \gamma), \text{ where } \gamma \sim \mathrm{Gumbel} (0,1)$.  In the sequential regime, we allow the rule application procedure (step 2, 3, 4 above) to be performed multiple times iteratively in $K$ rule application stages for each time-step $t$.

A pictorial demonstration of both rule application regimes can be found in Figure \ref{fig:seq_par}. We provide detailed algorithms for the sequential and parallel regimes in Appendix.

\vspace{-3mm}
\section{Experiments}
\vspace*{-1mm}

%\Ani{Make a figure describing all the tasks here.}
\iffalse

%\textbf{MNIST Transformations task}: NPS learns to perform any given transformations on MNIST digits by segregating each transformation into a separate rule.
\begin{figure*}
    \centering
    \includegraphics[width = 20cm]{figures/Exp_Demo.pdf}
    \caption{Demonstration of various experiments we use to prove the effectiveness of \modelname. \textbf{Arithmetic Task}: \modelname\ learns to solve mathematical equations by segregating each operation into a separate rule.  \textbf{Learning Action-Conditioned World Models}: NPS successfully predicts the outcomes of various actions by learning sparse interaction rules between objects (or entities) in the environment. \textbf{Learning Rules for Physical Reasoning Tasks}: NPS  succeeds in physical reasoning tasks in complex photo-realistic environments by learning physical rules that govern the dynamics of the objects.}
    \label{fig:experiments_demonstration}
\end{figure*}
\fi

\begin{wraptable}{r}{5.5cm}
   \scriptsize	
\centering 
\vspace*{-4mm}
\renewcommand{\arraystretch}{1.0}
\setlength{\tabcolsep}{1.3pt}
\caption{This table shows the segregation of rules for the MNIST Transformation task.  Each cell indicates the number of times the corresponding rule is used for the given operation. We can see that \modelname\ automatically and perfectly learns a separate rule for each operation.}\label{mnist_segretation}
 \begin{tabular}{|c|c|c|c|c|}
    \hline
    & Rule 1 & Rule 2 &  Rule 3 & Rule 4 \\
    \hline
    Translate Down & 5039 & 0 & 0 & 0 \\
    Translate Up & 0 & 4950 & 0 & 0 \\
     Rotate Right & 0 & 0 & 5030 & 0 \\
     Rotate Left & 0 & 0 & 0 & 4981 \\
    \hline

    \end{tabular}
    \vspace*{-2mm}
    \end{wraptable}

We demonstrate the effectiveness of \modelname\ on multiple tasks and compare to a comprehensive set of baselines. To show that \modelname\ can learn intuitive rules from the data generating distribution, we design a couple of simple toy experiments with well-defined discrete operations. Results show that \modelname\ can accurately recover each operation defined by the data and learn to represent each operation using a separate rule. We then move to a much more complicated and visually rich setting with abstract physical rules and show that factorization of knowledge into rules as offered by \modelname\ does scale up to such settings. We study and compare the parallel and sequential rule application procedures and try to understand the settings which favour each. We then evaluate the benefits of reusable, dynamic and sparse interactions as offered by \modelname\ in a wide variety of physical environments by comparing it against various baselines. We conduct ablation studies to assess the contribution of different components of \modelname. Here we briefly outline the tasks considered and direct the reader to the Appendix for full details on each task and details on hyperparameter settings.

\textbf{Discussion of baselines}. NPS is an interaction network, therefore we use other widely used interaction networks such as multihead attention and graph neural networks (\cite{goyal2019recurrent}, \cite{goyal2020object}, \cite{op3veerapaneni}, \cite{kipf2019contrastive}) for comparison. \cite{goyal2019recurrent} and \cite{goyal2020object} use an attention based interaction network to capture interactions between the slots, while \cite{op3veerapaneni} and \cite{kipf2019contrastive} use a GNN based interaction network. We also consider the recently introduced convolutional interaction network (CIN) \citep{qi2021learning} which captures dense pairwise interactions like GNN but uses a convolutional network instead of MLPs to better utilize spatial information. The proposed method, similar to other interaction networks, is agnostic to the encoder backbone used to encode the input image into slots, therefore we compare NPS to other interaction networks across a wide-variety of encoder backbones.

\vspace{-1mm}
\subsection{Learning intuitive rules with NPS: Toy Simulations}
%For a system to be called a production system, it should satisfy some prerequisites: 
%\begin{itemize}
%    \item It should be able to exhaustively extract all operations in the given environment and represent them as separate vectors or rules.   
    %\vspace{-0.3cm}
%    \item It should learn to apply the correct rules depending on the state of the variable/s.
%\end{itemize}
%we don't need "model" in front of the NPS. (I think). As NPS is already a proper noun.
% sure, makes sense

We designed a couple of simple tasks with well-defined discrete rules to show that \modelname\ can learn intuitive and interpretable rules. We also show the efficiency and effectiveness of the selection procedure (step 2 and step 3 in section \ref{section:computational_steps}) by comparing against a baseline with many more parameters.  Both tasks require a single modification of only one of the available entities, therefore the use of sequential or parallel rule application would not make a difference here since parallel rule application in which all-but-one slots select the null rule is similar to sequential rule application with 1 rule application step. %Therefore, we can use either rule application procedures. 
To simplify the presentation, we describe the setup for both tasks using the sequential rule application procedure.

\textbf{MNIST Transformation.} We test whether \modelname\  can learn simple rules for performing transformations on MNIST digits. We generate data with four transformations: \{Translate Up, Translate Down, Rotate Right, Rotate Left\}. We feed the input image ($\vX$) and the transformation ($\vo$) to be performed as a one-hot vector to the model. The detailed setup is described in Appendix. For this task, we evaluate whether NPS can learn to use a unique rule for each transformation.

We use 4 rules corresponding to the 4 transformations with the hope that the correct transformations are recovered. Indeed, we observe that {\bf \modelname\,  successfully learns to represent each transformation using a separate rule} as shown in Table \ref{mnist_segretation}. Our model achieves an MSE of 0.02. A visualization of the outputs from our model and further details can be found in Appendix \ref{mnist_appendix}. % For example, if we want to perform the operation \textit{Rotate Right}, we pick the rule vector that was assigned to the \textit{Rotate Right} operation during training and add that to the encoder output $\vZ$ for the corresponding digit. In this way, we can perform any combination of transformations on the digits using \modelname. 
%\fi
\begin{wrapfigure}{r}{0.5\textwidth}
\vspace{-1.5\baselineskip}

  \begin{center}
    \includegraphics[width=0.48\textwidth]{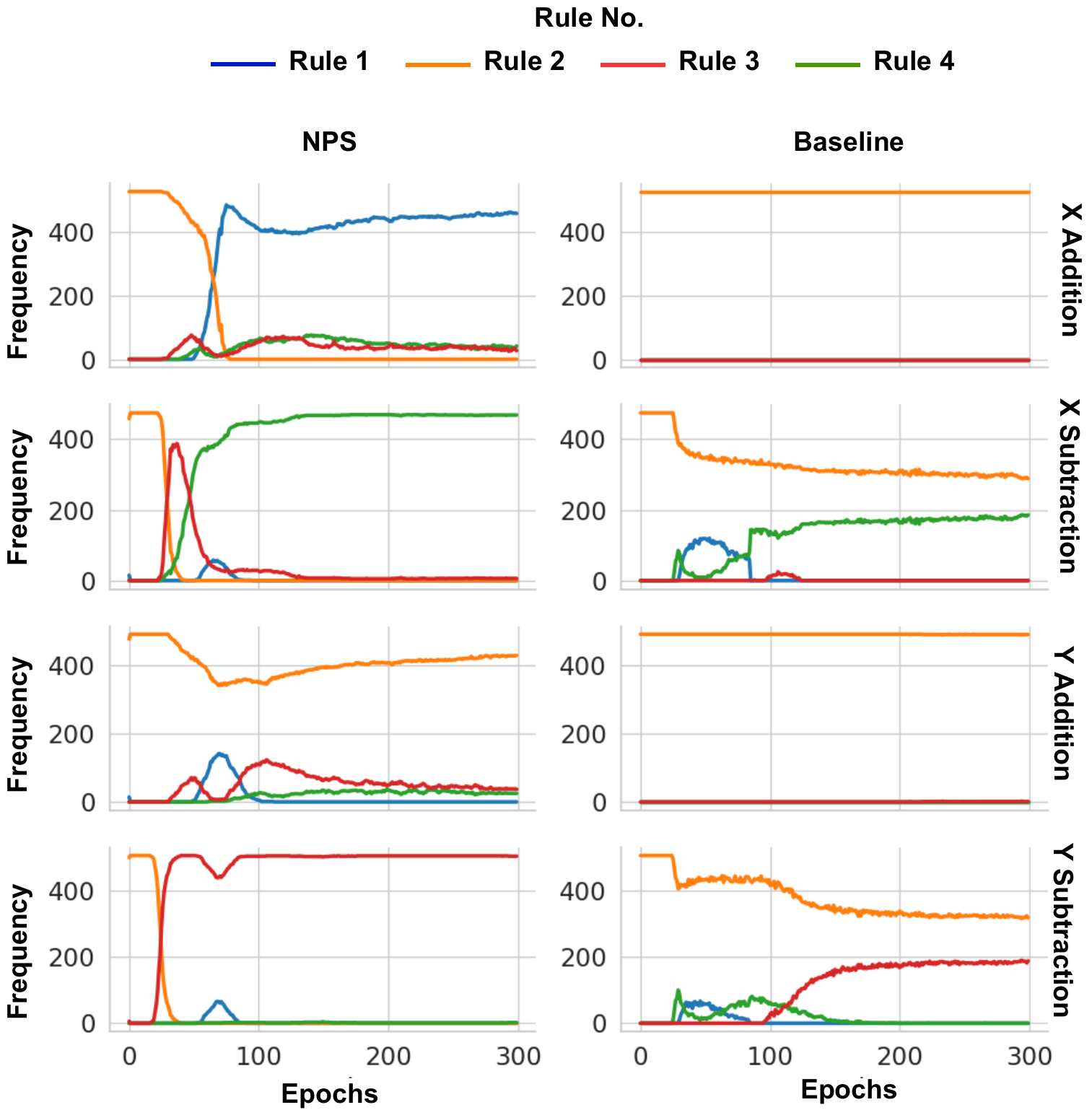}
  \end{center}
  \caption{\textbf{Coordinate Arithmetic Task.} Here, we compare \modelname\ to the baseline model in terms of segregation of rules as the training progresses. X-axis shows the epochs and Y-axis shows the frequency with which Rule $i$ is used for the given operation. We can see that \modelname\ disentangles the operations perfectly as training progresses with a unique rule specializing to every operation while the baseline model fails to do so.} \label{fig:coordinate_arithmetic_plot}
  \vspace{-2\baselineskip}
\end{wrapfigure}

\textbf{Coordinate Arithmetic Task.} The model is tasked with performing arithmetic operations on 2D coordinates. Given $(X_0, Y_0)$ and $(X_1, Y_1)$, we can apply the following operations: \{\textbf{X Addition}: $(X_r, Y_r) = (X_0 + X_1, Y_0)$, \textbf{X Subtraction}: $(X_r, Y_r) = (X_0 - X_1, Y_0)$, \textbf{Y Addition}: $(X_r, Y_r) = (X_0, Y_0 + Y_1)$, \textbf{Y Subtraction}: ${X_r, Y_r} = (X_0, Y_0 - Y_1)$\}, where $(X_r, Y_r)$ is the resultant coordinate.

In this task, the model is given 2 input coordinates $X = [(x_i, y_i), (x_j, y_j)]$ and the expected output coordinates $Y = [(\hat{x_i}, \hat{y_i}), (\hat{x_j}, \hat{y_j})]$ . The model is supposed to infer the correct rule to produce the correct output coordinates. During data collection, the true output is obtained by performing a random transformation on a randomly selected coordinate in $X$ (primary coordinate), taking another randomly selected coordinate from $X$ (contextual coordinate) into account. The detailed setup is described in Appendix \ref{appendix:coordinate_arithmetic}. We use an NPS model with 4 rules for this task. We use the the selection procedure in step 2 and step 3 of algorithm \ref{alg:neural_production_systems} to select the primary coordinate, contextual coordinate, and the rule. For the baseline we replace the selection procedure in NPS (i.e. step 2 and step 3 in algorithm \ref{alg:neural_production_systems}) with a routing MLP similar to \cite{swicth_transformers}.

This routing MLP has 3 heads (one each for selecting the primary coordinate, contextual coordinate, and the rule). The baseline has 4 times more parameters than NPS.  The final output is produced by the rule MLP which does not have access to the true output, hence the model cannot simply copy the true output to produce the actual output. Unlike the MNIST transformation task, we do not provide the operation to be performed as a one-hot vector input to the model, therefore it needs to infer the available operations from the data demonstrations. 

%Note that the expected output is only used by the model in the selection procedure for selecting the \{rule, primary coordinate, contextual coordinate\} triplet.

\begin{wraptable}{r}{5.5cm}
\vspace{-5mm}
\scriptsize	
\centering 
\renewcommand{\arraystretch}{1.3}
\setlength{\tabcolsep}{1.5pt}
\caption{This table shows segregation of rules when we use \modelname\ with 5 rules but the data generation distributions describes only 4 possible operations. We can see that only 4 rules get majorly utilized thus confirming that \modelname\ successfully recovers all possible operations described by the data. }\label{arithmetic_segretation}
% 5 rules
%0 {0: 360, 1: 99, 2: 45, 3: 13, 4: 0}
%1 {0: 0, 1: 57, 2: 15, 3: 99, 4: 335}
%2 {0: 0, 1: 482, 2: 0, 3: 1, 4: 0}
%3 {0: 0, 1: 39, 2: 453, 3: 2, 4: 0}

 \begin{tabular}{|c|c|c|c|c|c|}
    \hline
    & Rule 1 & Rule 2 &  Rule 3 & Rule 4 & Rule 5 \\
    \hline
    X Addition & 360 & 99 & 45 & 13 & 0 \\
    X Subtraction & 0 & 482 & 0 & 1 & 0 \\
    Y Subtraction & 0 & 39 & 453 & 2 & 0 \\
    Y Addition & 0 & 57 & 15 & 99 & 335 \\
    \hline
    \end{tabular}
  \vspace{-3\baselineskip}
\end{wraptable} 

We show the segregation of rules for \modelname\ and the baseline in Figure \ref{fig:coordinate_arithmetic_plot}. \textbf{We can see that \modelname\ learns to use a unique rule for each operation while the baseline struggles to disentangle the underlying operations properly}. \textbf{NPS also outperforms the baseline in terms of MSE} achieving an MSE of \g{0.01}{0.001} while the baseline achieves an MSE of \g{0.04}{0.008}.  To further confirm that \modelname\ learns all the available operations correctly from raw data demonstrations, we use an NPS model with 5 rules. \textbf{We expect that in this case NPS should utilize only 4 rules since the data describes only 4 unique operations and indeed we observe that NPS ends up mostly utilizing 4 of the available 5 rules} as shown in Table \ref{arithmetic_segretation}.

%\vspace{-2mm}
\subsection{Parallel vs Sequential Rule Application}\label{section:exp_par_seq}
%\vspace{-2mm}
We compare the parallel and sequential rule application procedures, to understand the settings that favour one or the other, over two tasks: (1) Bouncing Balls, (2) Shapes Stack. We use the term \textsc{PNPS} to refer to parallel rule application and \textsc{SNPS} to refer to sequential rule application.

\vspace{-2mm}

\begin{wraptable}{r}{5.5cm}
\vspace{-4mm}
\centering
\scriptsize	
\centering 
\renewcommand{\arraystretch}{1.3}
\setlength{\tabcolsep}{1.5pt}
\begin{tabular}{|c|c|c|}
\hline
        Model Name & Test &   Transfer   \\
        \hline
        RPIN (\cite{qi2021learning}) & \g{1.254}{0.008} & \g{6.377}{0.325} \\
        \hline
        PNPS  &  \highlight{\g{1.250}{0.007}} & \highlight{\g{5.411}{0.45}} \\
        SNPS  & \g{1.68}{0.02} & \g{5.80}{0.15} \\
        \hline
\end{tabular}
\caption{Prediction error of the compared models on the shapes stack dataset (lower is better) for the test as well as transfer setting. In the test setting the number of rollout steps $\vt$ is set to 15 and in the transfer setting it is set to 30. We can see that PNPS outperforms the RPIN baseline in both the test and transfer setting while SNPS fails to do so. Results across 15 seeds.}  
\label{tab:ss_table_1}
\vspace{-2\baselineskip}
\end{wraptable}

\paragraph{Shapes Stack.} We use the shapes stack dataset introduced by \citet{shapestack}. This dataset consists of objects stacked on top of each other as shown in Figure \ref{fig:ss_demo}. These objects fall under the influence of gravity. For our experiments, We follow the same setup as \cite{qi2021learning}. In this task, given the first frame, the model is tasked with predicting the object bounding boxes for the next $\vt$ timesteps. %In the test setting, $\vt$ is set to 15. In the transfer setting, $\vt$ is set to 30. 
The first frame is encoded using a convolutional network followed by RoIPooling (\cite{fast-rcnn}) to extract object-centric visual features. The object-centric features are then passed to the dynamics model to predict object bounding boxes of the next $\vt$ steps. \cite{qi2021learning} propose a Region Proposal Interaction Network (RPIN) to solve this task. The dynamics model in RPIN consists of an Interaction Network proposed in \cite{battagliainteraction}. To better utilize spatial information, \cite{qi2021learning} propose an extension of the interaction operators in interaction net to operate on 3D tensors. This is achieved by replacing the MLP operations in the original interaction networks with convolutions. They call this new network Convolutional Interaction Network (CIN). For the proposed model, we replace this CIN in RPIN by NPS. To ensure a fair comparison to CIN, we use CNNs to represent rules in NPS instead of MLPs. CIN captures all pairwise interactions between objects using a convolutional network. In NPS, we capture sparse interactions (contextual sparsity) as compared to dense pairwise interactions captured by CIN. Also, in NPS we update only a few subset of slots per step instead of all slots (application sparsity).

We consider two evaluation settings. (1) \textbf{Test setting}: The number of rollout timesteps is same as that seen during training (i.e. $\vt = 15$); (2) \textbf{Transfer Setting}: The number of rollout timesteps is higher than that seen during training (i.e. $\vt = 30$).

We present our results on the shapes stack dataset in Table \ref{tab:ss_table_1}. We can see that both \textsc{PNPS} and \textsc{SNPS} outperform the baseline RPIN in the transfer setting, while only PNPS outperforms the baseline in the test setting and SNPS fails to do so. We can see that \textsc{PNPS} outperforms \textsc{SNPS}. We attribute this to the reduced \textit{application sparsity} with \textsc{PNPS}, i.e., it is more likely that the state of a slot gets updated in \textsc{PNPS} as compared to \textsc{SNPS}. For instance, consider an NPS model with $\vN$ uniformly chosen rules and $\vM$ slots. The probability that the state of a slot gets updated in \textsc{PNPS} is $P_{PNPS} = \vN - 1 / \vN$ (since 1 rule is the null rule), while the same probability for \textsc{SNPS} is $P_{SNPS} = 1 / \vM$ (since only 1 slot gets updated per rule application step).

For this task, we run both \textsc{PNPS} and \textsc{SNPS} for $\vN = \{1,2,4,6\}$ rules and $\vM = 3$. For any given $\vN$, we observe that $P_{PNPS} > P_{SNPS}$. Even when we have multiple rule application steps in \textsc{SNPS}, it might end up selecting the same slot to be updated in more than one of these steps. We report the best performance obtained for \textsc{PNPS} and \textsc{SNPS} across all $\vN$, which is $\vN = \{2 + 1 \quad \text{Null Rule}\}$ for \textsc{PNPS} and $\vN = 4$ for \textsc{SNPS}, in Table \ref{tab:ss_table_1}. Shapes stack is a dataset that would prefer a model with less application sparsity since all the objects are tightly bound to each other (objects are placed on top of each other), therefore all objects spend the majority of their time interacting with the objects directly above or below them. We attribute the higher performance of \textsc{PNPS} compared to RPIN to the higher contextual sparsity of \textsc{PNPS}. Each example in the shapes stack task consists of 3 objects. Even though the blocks are tightly bound to each other, each block is only affected by the objects it is in direct contact with. For example, the top-most object is only affected by the object directly below it. The contextual sparsity offered by \textsc{PNPS} is a strong inductive bias to model such sparse interactions while RPIN models all pairwise interactions between the objects. Figure \ref{fig:ss_demo} shows an intuitive illustration of the \textsc{PNPS} model for the shapes stack dataset. In the figure, \textit{Rule 2} actually refers to the Null Rule, while \textit{Rule 1} refers to all the other non-null rules. The bottom-most block picks the Null Rule most times, as the bottom-most block generally does not move. 
\begin{wrapfigure}{r}{0.50\textwidth}
\vspace{-1.0\baselineskip}
  \includegraphics[width=0.45\textwidth]{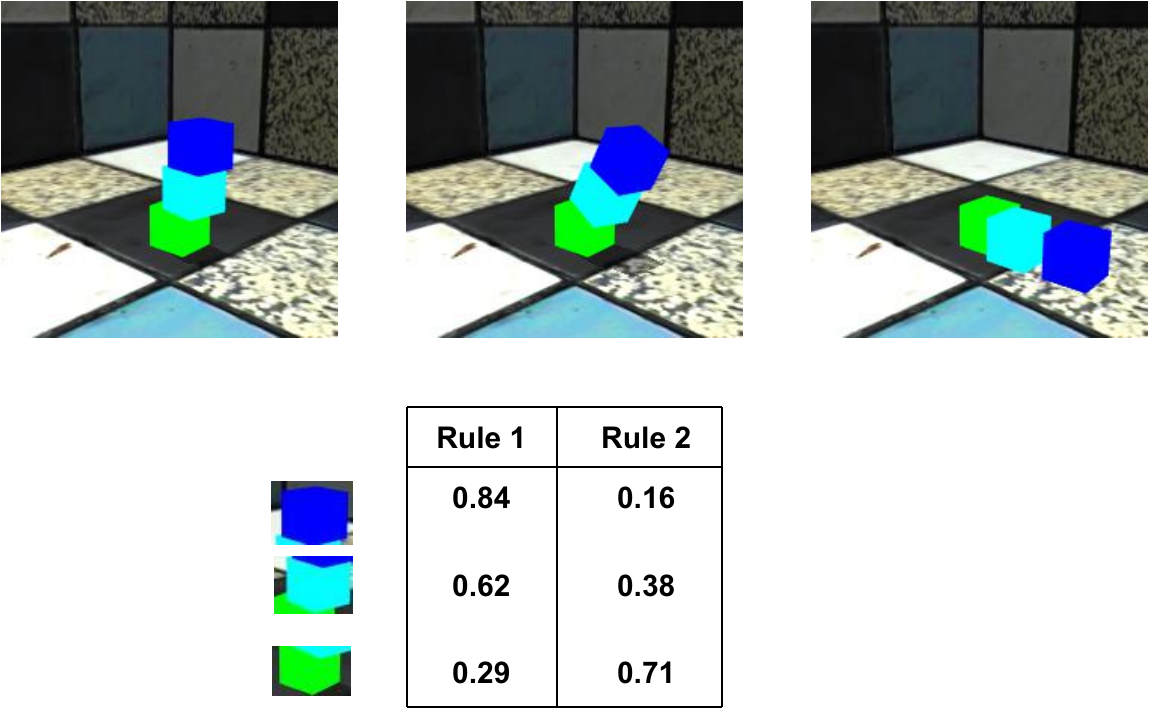}
  \caption{Here we show the rule selection statistics from the proposed model for all entities in the shapes stack dataset across all examples. Each example contains 3 entities as shown above. Each cell in the table shows the probability with which the given rule is triggered for the corresponding entity. We can see that the bottom-most entity triggers rule 2 most of the time while the other 2 entities trigger rule 1 most often. This is quite intuitive as, for most examples, the bottom-most entity remains static and does not move at all while the upper entities fall. Therefore, rule 2 captures information which is relevant to static entities, while rule 1 captures physical rules relevant to the interactions and motion of the upper entities.} \label{fig:ss_demo}
\vspace{-2.5\baselineskip}
\end{wrapfigure}
\vspace{-5mm}
\paragraph{Bouncing Balls.}

We consider a bouncing-balls environment in which multiple balls move with billiard-ball dynamics. We validate our model on a colored version of this dataset. This is a next-step prediction task in which the model is tasked with predicting the final binary mask of each ball.  We compare the following methods: (a) SCOFF \citep{goyal2020object}:  factorization of knowledge in terms of slots (object properties) and schemata, the latter capturing object dynamics; (b) SCOFF++: we extend SCOFF by using the idea of iterative competition as proposed in slot attention (SA) \citep{locatello2020object}; SCOFF + PNPS/SNPS: We replace pairwise slot-to-slot interaction in SCOFF++ with parallel or sequential rule application. For comparing different methods, we use the Adjusted Rand Index or ARI~\citep{rand1971objective}. To investigate how the factorization  in the form of rules allows for extrapolating knowledge from fewer to more objects, we increase the number of objects from 4 during training to 6-8 during testing.

We present the results of our experiments in Table \ref{tab:bb_table_1}. Contrary to the shapes stack task, we see that \textsc{SNPS} outperforms \textsc{PNPS} for the bouncing balls task. The balls are not tightly bound together into a single tower as in the shapes stack. Most of the time, a single ball follows its own dynamics, only occasionally interacting with another ball. Rules in NPS capture interaction dynamics between entities, hence they would only be required to change the state of an entity when it interacts with another entity. In the case of bouncing balls, this interaction takes place through a collision between multiple balls. Since for a single ball, such collisions are rare, \textsc{SNPS}, which has higher application sparsity (less probability of modifying the state of an entity), performs better as compared to \textsc{PNPS} (lower application sparsity). Also note that, \textsc{SNPS} has the ability to compose multiple rules together by virtue of having multiple rule application stages. A visualization of the rule and entity selections by the proposed algorithm can be found in Appendix Figure \ref{fig:bb_vis}.

Given the analysis in this section, we can conclude that \textsc{PNPS} is expected to work better when interactions among entities are more frequent while \textsc{SNPS} is expected to work better when interactions are rare and most of the time, each entity follows its own dynamics. Note that, for both \textsc{SNPS} and \textsc{PNPS}, the rule application considers only 1 other entity as context. Therefore, both approaches have equal \textit{contextual sparsity} while the baselines that we consider (SCOFF and RPIN) capture dense pairwise interactions. We discuss the benefits of \textit{contextual sparsity} in more detail in the next section. More details regarding our setup for the above experiments can be found in Appendix.

%\vspace{-3mm}
\subsection{Benefits of Sparse Interactions Offered by NPS}\label{section:sparse} %\label{sec:action_condition}
%\vspace{-2mm}

In \modelname, one can view the computational graph as a dynamically constructed GNN resulting from applying dynamically selected rules, where the states of the slots are represented on the different nodes of the graph, and different rules dynamically instantiate an hyper-edge between a set of slots (the primary slot and the contextual slot). It is important to emphasize that the topology of the graph induced in \modelname\ is dynamic and sparse (only a few nodes affected), while in most GNNs the topology is fixed and dense (all nodes affected). In this section, through a thorough set of experiments, we show that learning sparse and dynamic interactions using \modelname\ indeed works better for the problems we consider than learning dense interactions using GNNs. We consider two types of tasks: (1) Learning Action Conditioned World Models (2) Physical Reasoning. We use \textsc{SNPS} for all these experiments since in the environments that we consider here, interactions among entities are rare.
\begin{wraptable}{r}{5.5cm}
\centering
\scriptsize	
\centering 
\renewcommand{\arraystretch}{1.3}
\setlength{\tabcolsep}{1.5pt}

\begin{tabular}{|c|c|c|}
\hline
        Model Name & Test  & Transfer   \\
        \hline
        SCOFF & 0.28 & 0.15 \\
        SCOFF++ &0.8437 & 0.2632  \\
        \hline
        PNPS (10 Rules+1 Null Rule) & 0.7813  & 0.1997 \\
        SNPS (10 Rules) &  \highlight{0.8518} & \highlight{0.3553} \\
        \hline
\end{tabular}
\caption{Here we show the ARI achieved by the models on the bouncing balls dataset (higher is better). We can see that SNPS outperforms SCOFF and SCOFF++ while PNPS has a poor performance in this task. Results average across 2 seeds.}  \label{tab:bb_table_1}

    \vspace{-4.0\baselineskip}
\end{wraptable}

\paragraph{Learning Action-Conditioned World Models.} For learning action-conditioned world models, we follow the same experimental setup as \citet{kipf2019contrastive}. Therefore, all the tasks in this section are next-$K$ step ($K = \{1, 5, 10\}$) prediction tasks, given the intermediate actions, and with the predictions being performed in the latent space. We use the Hits at Rank 1 (H@1) metrics described by \citet{kipf2019contrastive} for evaluation. H@1 is 1 for a particular example if the predicted state representation is nearest to the encoded true observation and 0 otherwise. We report the average of this score over the test set (higher is better). %For a description of MRR and the results on the MRR metric, we ask the reader to refer to the appendix. The proposed \modelname\  is  used as a drop-in replacement for the GNN used in the C-SWM model. We find that capturing sparse interactions using NPS is better suited for learning action-conditioned world models compared to GNNs in a wide range of settings as described below.

%\textbf{Mean Reciprocal Rank (MRR)}: This is defined as the average inverse rank, i.e, MRR $= \frac{1}{N} \sum_{n=1}^N \frac{1}{\text{rank}_n}$ where $\text{rank}_n$ is the rank of the $n^{th}$ sample of the test set where ranking is done over all reference state representations. Here also, higher MRR indicates better performance. 

\paragraph{Physics Environment.} The physics environment \citep{ke2021systematic} simulates a simple physical world. It consists of blocks of unique but unknown weights. The dynamics for the interaction between blocks is that the movement of heavier blocks pushes lighter blocks on their path. This rule creates an acyclic causal graph between the blocks. For an accurate world model, the learner needs to infer the correct weights through demonstrations. Interactions in this environment are sparse and only involve two blocks at a time, therefore we expect \modelname\ to outperform dense architectures like GNNs. This environment is demonstrated in Appendix Fig \ref{fig:weighted_blocks}.

We follow the same setup as \cite{kipf2019contrastive}. We use their C-SWM model as baseline. For the proposed model, we only replace the GNN from C-SWM by \modelname. GNNs generally share parameters across edges, but in \modelname\ each rule has separate parameters. For a fair comparison to GNN, we use an \modelname\ model with 1 rule. Note that this setting is still different from GNNs as in GNNs at each step every slot is updated by instantiating edges between all pairs of slots, while in \modelname\ an edge is dynamically instantiated between a single pair of slots and only the state of the selected slot (i.e., primary slot) gets updated.

%The agent performs stochastic interventions (actions) in the environment to infer the weights of the blocks. Each intervention makes a block move in any of the 4 available directions (left, right, up, and down). When an intervened block $A$ with weight $W_A$ comes into contact with another block $B$ with weight $W_B$, the block $B$ may get pushed if $W_B < W_A$ else $B$ will remain still. Hence, any interaction in this environment involves only 2 blocks (i.e., 2 entities) and the other blocks are not affected. Therefore, modelling the interactions between all blocks for every intervention, as is generally done using GNNs, may be wasteful. NPS is particularly well suited for this task as  any rule application takes into account only a subset of entities, hence considering interactions between those blocks only. Note that the interactions in this environment are not symmetrical and NPS can handle such relations. For example, consider a set of 3 blocks: $\{A_0, A_1, A_2\}$. If an intervention leads to $A_1$ pushing $A_0$, then NPS would apply the rule to the slot representing entity $A_0$. Since the movement of $A_0$ would depend on whether $A_0$ is heavier or lighter than $A_1$, NPS would also select the slot representing entity $A_1$ as a contextual slot and take it into account while applying the rule to $A_0$. Therefore, NPS can represent sparse and directed rules, which as we show, is more useful in this environment then learning dense and undirected relationships (or commutative operations). 

\begin{figure*}[th]
    \centering
    \vspace*{-7mm}
    \subfigure[\textbf{Physics Env}]{\includegraphics[width=0.45\textwidth]{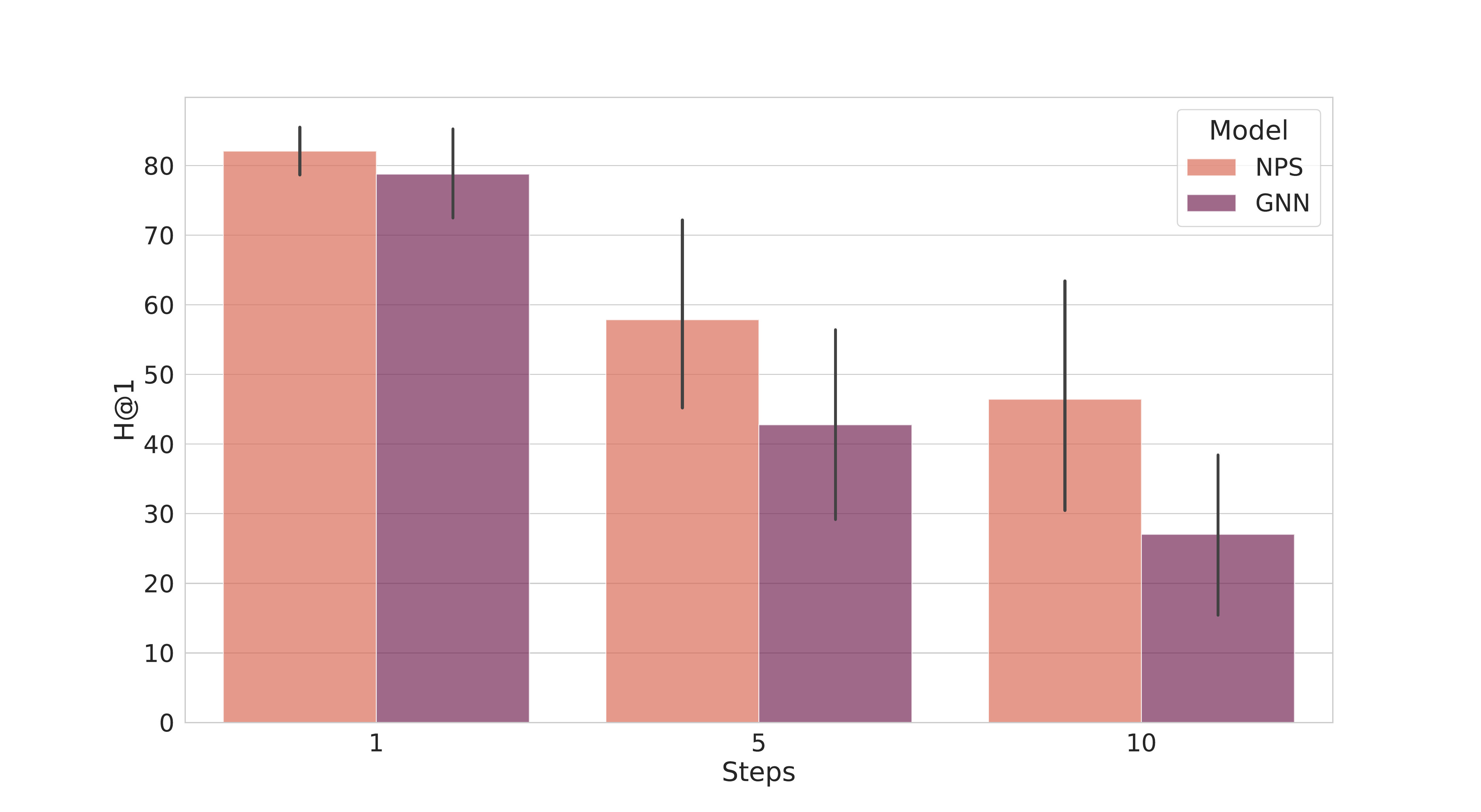}} 
    %\subfigure[\textbf{Physics Env}]{\includegraphics[width=0.32\textwidth]{figures/Physics/physics_plot_3.pdf}} 
    \subfigure[\textbf{Atari Games}]{\includegraphics[width=0.45\textwidth]{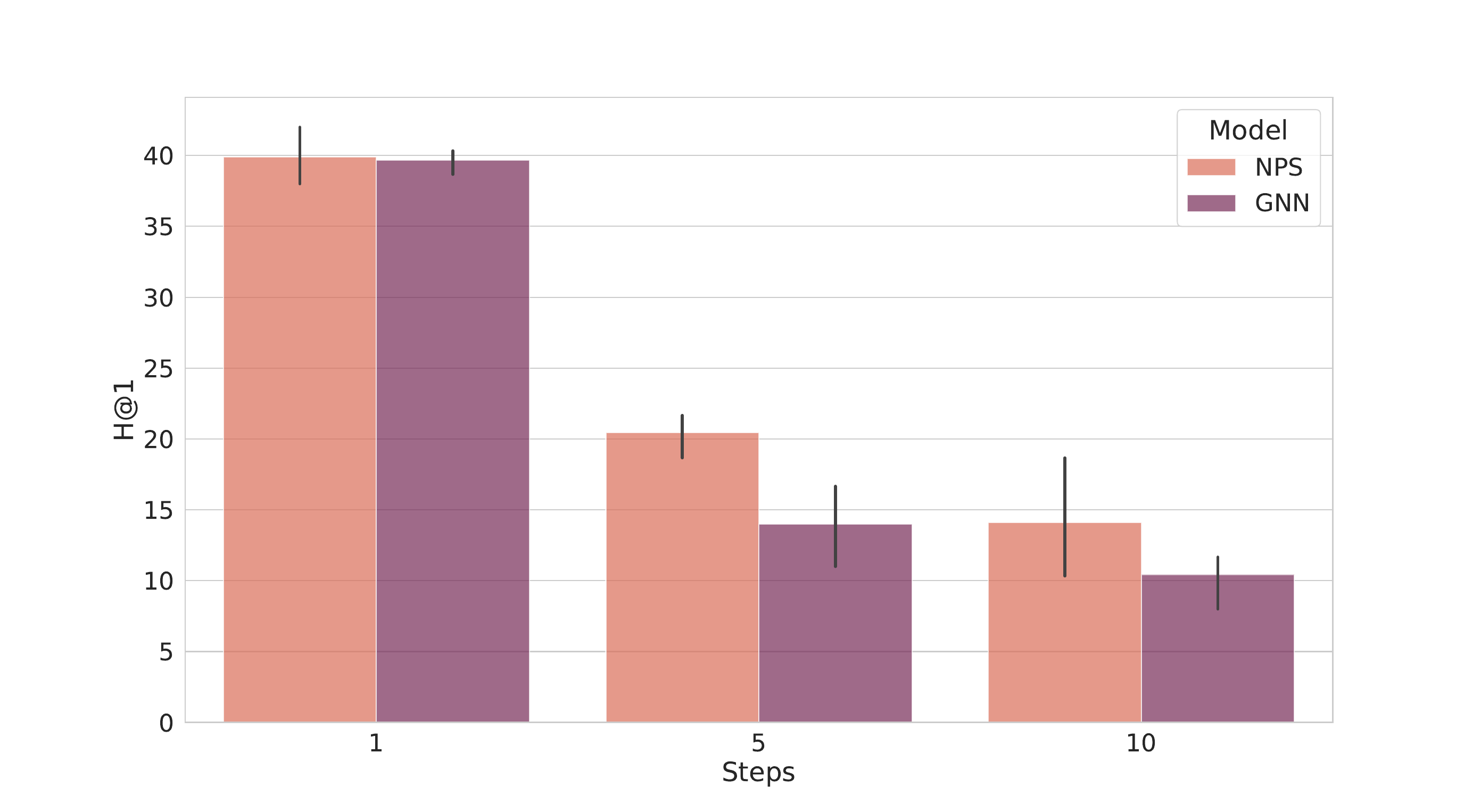}}\par
    \vspace{-2mm}
    \caption{ \textbf{Action-Conditioned World Models}, with number of future  steps to be predicted for the world-model on the horizontal axes. (a) Here we show a comparison between GNNs and the proposed \modelname\ on the physics environment using the H@1 metric (higher is better). %(b) Effect of number of rule application steps used in NPS when using a single rule in the physics environment. 
    (b) Comparison of average H@1 scores across 5 Atari games for the proposed model \modelname\ and GNN.
    }
    \label{fig:gnn_experiments}
    \vspace*{-2mm}
\end{figure*}

%For training, we follow the same setup as \cite{kipf2019contrastive}. We use the C-SWM model as baseline. For the proposed model, we only replace the GNN from C-SWM by the proposed \modelname. Graph neural networks are generally implemented by sharing parameters across edges, we can also implement them using separate parameters per edge but that would break their equivariance (as we add more entities). For the proposed \modelname\ model we use separate parameters per rule. Therefore, for a fair comparison to the GNN baseline in terms of number of parameters, we run \modelname\ with a single rule. Note that this setting is still different from GNNs as in GNNs at each step every slot is updated by instantiating edges between all pairs of slots, while in \modelname\ an edge is dynamically instantiated between a single pair of slots and only the state of the selected slot gets updated.

%For training, we follow the same setup as \cite{kipf2019contrastive}. To summarize, we use a convolutional encoder to encode the image into a set of entities, each representing a particular block. We then use a transition model for next-step prediction in the latent space. This transition model can either be a GNN or \modelname. The details of the loss can be found in the appendix section \ref{contrastiveloss}.

The results of our experiments are presented in Figure \ref{fig:gnn_experiments}(a). We can see that \modelname\ outperforms GNNs for all rollouts. Multi-step settings are more difficult to model as errors may get compounded over time steps. The sparsity of \modelname\ (only a single slot affected per step) reduces compounding of errors and enhances symmetry-breaking in the assignment of transformations to rules, while in the case of GNNs, since all entities are affected per step, there is a higher possibility of errors getting compounded. We can see that even with a single rule, we significantly outperform GNNs thus proving the effectiveness of dynamically instantiating edges between entities. %Using a single rule is appropriate for the physics environment as this environment affords only a single rule, i.e., heavier blocks push lighter blocks. For more compositional environments such as those explored in the next section, we find that having multiple rules improves performance. %We can see that generally models with 5 rules perform better than models with 10 rules in this case. This can be attributed to the simple nature of the physics environment hence having as many as 10 rules may not be required in this environment. 

\paragraph{Atari Games.} We also test the proposed model in the more complicated setting of Atari. Atari games also have sparse interactions between entities. For instance, in Pong, any interaction involves only 2 entities: (1) paddle and ball or (2) ball and the wall. Therefore, we expect sparse interactions captured by \modelname\ to outperform GNNs here as well.

We follow the same setup as for the physics environment described in the previous section. We present the results for the Atari experiments in Figure \ref{fig:gnn_experiments}(b), showing the average H@1 score across 5 games: Pong, Space Invaders, Freeway, Breakout, and QBert. As expected, we can see that the proposed model achieves a higher score than the GNN-based C-SWM. The results for the Atari experiments reinforce the claim that \modelname\ is especially good at learning sparse interactions.
\paragraph{Learning Rules for Physical Reasoning.}
%\vspace{-2mm}

%Physical reasoning includes modelling the dynamics of various objects and how these objects interact with other objects. An optimal model would use  use the same knowledge about the dynamics and how collisions affect the states of different interacting objects  across all objects, yet will disentangle the different objects and their meaningful attributes to represent their individual states. 

To show the effectiveness of the proposed approach for  physical reasoning tasks, we evaluate \modelname\ on another dataset: Sprites-MOT \citep{spmot}. The Sprites-MOT dataset was introduced by \citet{spmot}. The dataset contains a set of moving objects of various shapes. This dataset aims to test whether a model can handle occlusions correctly. Each frame has consistent bounding boxes which may cause the objects to appear or disappear from the scene. A model which performs well should be able to track the motion of all objects irrespective of whether they are occluded or not. We follow the same setup as \cite{weis2020unmasking}. We use the OP3 model \citep{op3veerapaneni} as our baseline. To test the proposed model, we replace the GNN-based transition model in OP3 with the proposed \modelname. %In this setup, the model predicts a separate mask for each object in the ground truth frames.

\begin{wraptable}{r}{5.5cm}
\centering
\scriptsize	
\centering 
\renewcommand{\arraystretch}{1.6}
\setlength{\tabcolsep}{2.0pt}

\begin{tabular}{|c|c|c|}
\hline
        Model & MOTA $\uparrow$  & MOTP $\uparrow$   \\
        \hline
        OP3 & \g{89.1}{5.1} & \g{78.4}{2.4} \\
        NPS & \highlight{\g{90.72}{5.15}} & \highlight{\g{79.91}{0.3} } \\
        \hline
        
\end{tabular}
\caption{\textbf{Sprites-MOT}. Comparison between the proposed \modelname\ and the baseline OP3 using the MOTA and MOTP metrics on the sprites-MOT dataset ($\uparrow$: higher is better). Average over 3 random seeds.}  \label{tab:spmot_1}

    \vspace{-2.0\baselineskip}

\end{wraptable}
We use the same evaluation protocol as followed by \cite{weis2020unmasking} which is based on the MOT (Multi-object tracking) challenge \citep{DBLP:journals/corr/MilanL0RS16}. The results on the MOTA and MOTP metrics for this task are presented in Table \ref{tab:spmot_1}. The results on the other metrics are presented in appendix Table \ref{tab:spmot}.  We ask the reader to refer to appendix \ref{appendix:spmot_metrics} for more details about these metrics. We can see that for almost all metrics, \modelname\ outperforms the OP3 baseline. Although this dataset does not contain physical interactions between the objects, sparse rule application should still be useful in dealing with occlusions. At any time step, only a single object is affected by occlusions i.e., it may get occluded due to another object or due to a prespecified bounding box, while the other objects follow their default dynamics. Therefore, a rule should be applied to only the object (or entity) affected (i.e., not visible) due to occlusion and may take into account any other object or entity that is responsible for the occlusion.

\vspace{-3mm}
\section{Discussion and Conclusion}
\vspace{-1mm}

%\paragraph{Deep Learning and Knowledge Factorization.} In learning to model a visual environment,  many solutions could be found which fit the data well but will not generalize in a systematic way to different settings. To achieve systematic generalization, the knowledge about individual entities (in the form of slots) and sparse rules should be factored from each other.  This will allow any rule to be applied to multiple slots or  many rules can be applied to the same slot when appropriate. Although the above factorization may seem naturally desirable, it is not built into the neural network architectures: they lump together (for each artificial neuron) both the specification of a rule (via the synaptic weights of the neuron)  and the values of the attributes of entities (via the neural activity). We conjecture -- and experiments with \modelname\ show -- that  attention mechanisms offer even more flexibility in factorizing sparse rules and slots and combine them in arbitrary ways demanded by \emph{context and the attributes of entities}.

%\paragraph{Resurrecting Production Systems.} 
%\paragraph{
%{\bf Looking Backward.} Production systems were one of the first AI research attempts to model cognitive behaviour and form the basis of many existing models of cognition. However, in traditional symbolic AI, both the key entities and the rules that operated on the entities were given.  
For AI agents such as robots trying to make sense of their environment, the only observables are low-level variables like pixels in images. To generalize well, an agent must induce high-level entities as well as discover and disentangle the rules that govern how these entities actually interact with each other. Here we have focused on perceptual inference problems and proposed \modelname, a neural instantiation of production systems by introducing an important  inductive bias in the architecture following the proposals of \citet{marcus2003algebraic, bengio2017consciousness, goyal2020inductive, ke2021systematic}.
\vspace{3mm}

%\paragraph{
{\bf Limitations \& Looking Forward.} Our experiments %learning action-conditioned world models and extrapolation of knowledge in the form of learned rules in video prediction
highlight the advantages brought by the factorization of knowledge into a small set of entities and sparse sequentially applied rules. Immediate future work would investigate how to take advantage of these inductive biases for more complex physical environments \citep{ahmed2020causalworld} and novel planning methods, which might be more sample efficient than standard ones \citep{schrittwieser2020mastering}. 

We also find that Sequential and Parallel NPS have different properties suited towards different domains. Future work should explore how to effectively combine these two approaches. We discuss this in more detail in Appendix section \ref{parvsseq}. %Humans seem to exploit the inductive bias in the sparsity of the rules and that reasoning about the application of these rules in an abstract space can be very efficient.\\
%For such problems, exploration becomes a bottleneck but we believe using rules as a source of \textit{behavioural priors} can drive the necessary exploration \citep{goyal2019infobot,tirumala2020behavior, badia2020agent57}.
%\paragraph{
%{\bf Social Impact.} The authors do not foresee negative social impact of this work beyond that which could arise from general improvements in ML.
% Do we need social impact ?
% YB: yes

%Connectionist networks are good at performing pattern matching, especially when there is no perfect match and the aim is to find the best partial match. This falsifies any strong claim that connectionist systems using distributed representations could not possibly implement symbol processing.  \Ani{Talk about this.}
%Much larger passive memory, and much smaller active memory. Connect to inductive biases paper.

\section{Acknowledgements}
The authors would also like to thank Alex Lamb, Stefan Bauer, Nicolas Chapados,  Danilo Rezende, Matthew Botvinick and Kelsey Allen for brainstorming sessions. We are also thankful to Dianbo Liu, Damjan Kalajdzievski and Osama Ahmed for proofreading. We would like to thank Samsung Electronics Co. Ltd. and CIFAR for funding this research. We would also like to thank Google for providing Google cloud credits used in this work.

\bibliography{neurips_2021}
\bibliographystyle{plainnat}

\appendix
\onecolumn

\section*{\Large Appendix}

\begin{algorithm}
    \caption{ Sequential Neural Production System model}
   \label{alg:neural_production_systems}
   \begin{algorithmic}%[1]
   \footnotesize
   %\For{}
   \item[]
   
   \STATE{\bfseries \itshape Input:} Input sequence $\{ \vx^1, \ldots, \vx^t, \ldots,  \vx^T\}$, set of embeddings describing the rules $\Vec{\vR_i}$, and set of MLPs $(MLP_{i})$ corresponding to each rule $\vR_{1 \ldots N}$. Hyper-parameters specific to \modelname\ are the number of stages $K$, the number of slots $M$, and the
   number of rules $N$. $W^k$, $W^q$, $\widetilde{W}^k$, and $\widetilde{W}^q$ are learnable weights.

    \BlankLine 
    \item[]
    \For{each input element $\vx^t$ with $t \gets 1$ to $T$}  {
    
    \vspace{2mm}
    \STATE{{\bfseries \itshape Step 1:} \itshape Update or infer the entity state in each slot $j$, $\vV_j^{t,0}$, from
    the previous state, $\vV_j^{t-1,K}$ and the current input $\vx_t$.}
   %\STATE{\bfseries \itshape Step 1: Representation of  different entities  $\{ \vV_1^{t}, \ldots, \ldots, \vV_M^{t}\}$ at a particular time-step $t$. Function of the input at time step $t$ as well as representation of entities at time-step $t-1$ i.e., $\{ \vV_1^{t-1}, \ldots, \ldots, \vV_M^{t-1}\}$ } 
    %\vspace{2mm}
   \item[]
   %\STATE{\bfseries \itshape Repeat Procedure consisting of slot-rule pair selection to identify primary slot (Step 2), seeking information from other slots i.e., selecting contextual slot (Step 3), and applying the selected rule to update the state of primary slot  (Step 4) for K stages. } \\
   \vspace{2mm}
   \For{each stage $h \gets 0$ to $K-1$}  {                  
        \item[]
   \STATE{{\bfseries \itshape Step 2:} \itshape Select \{rule, primary slot\} pair}
   %\vW^q
  \STATE \bull $\vk_{i} = \Vec{\vR}_{i} \vW^k  \quad \forall i \in \{1, \ldots, N\}$
  \STATE \bull $\vq_j = \vV_{j}^{t,h} \vW^q \quad \forall j \in \{1, \ldots, M\}$
  \STATE \bull $r,p = \mathrm{argmax}_{i, j}\left(\vq_{j}\vk_i + \gamma \right)$
  \STATE $~~~~~~~~~~~~~~~\text{ where } \gamma \sim \mathrm{Gumbel} (0,1)$
  
  \item[]
   \STATE{{\bfseries \itshape Step 3:} \itshape Select contextual slot }
   
   \STATE \bull $\vq_{r, p} = V_{p}^{t, h}\widetilde{\vW}^q $
   \STATE \bull $\vk_j = \vV_{j}^{t,h} \widetilde{\vW}^k \quad \forall j \in \{1, \ldots, M\}$
   \STATE \bull $c = \mathrm{argmax}_j \left(\vq_{r, p} \vk_j + \gamma\right)$
   \STATE $~~~~~~~~~~~~~~~\text{ where } \gamma \sim \mathrm{Gumbel} (0,1)$
   
   \item[]
   
   \STATE{{\bfseries \itshape Step 4:} \itshape Apply selected rule to primary slot conditioned on contextual slot}
   
   \STATE \bull $\widetilde{\vR} = \mathrm{MLP}_{r}(\mathrm{Concatenate}([\vV_{p}^{t,h}, \vV_{c}^{t,h}]))$
   
   \STATE \bull $\vV_{p}^{t,h+1} = \vV_{p}^{t,h} + \widetilde{\vR}$
  
    }
   }
   
   %\item[]
   
   %\STATE{\bfseries \itshape Step 4: Apply the selected rule $\vR_{\hat{\vi}}$ to the selected entity $\vV_{\hat{\vj}}^{t}$ by taking into account relevant entity $\vV_{\hat{\vr}}^{t}$ }
   
   %\STATE \bull $\widetilde{\vR} = \mathrm{MLP}_{\hat{\vi}}(\mathrm{Concatenate}([\vV_{\hat{\vj}}^{t}, \vV_{\hat{\vr}}^{t}]))$
   
   %\STATE \bull $\vV_{\hat{\vj}}^{t} = \vV_{\hat{\vj}}^{t} + \widetilde{\vR}$
   %\EndFor
   
   %\item[]
   %\STATE{\bfseries \itshape Step 5: Repeat Procedure consisting of entity-rule pair selection (Step 2), seeking information from other relevant entity  (Step 3), and applying the selected rule  (Step 4).}

\end{algorithmic}
\end{algorithm}

\begin{algorithm}[t!]
    \caption{ Parallel Neural Production System model}
   \label{alg:par_neural_production_systems}
   \begin{algorithmic}%[1]
   \footnotesize
   %\For{}
   \item[]
   
   \STATE{\bfseries \itshape Input:} Input sequence $\{ \vx^1, \ldots, \vx^t, \ldots,  \vx^T\}$, set of embeddings describing the applicable rules $\Vec{\vR_i}$, set of MLPs $(MLP_{i})$ corresponding to each rule $\vR_{1 \ldots N}$,and an embedding vector corresponding to the Null Rule $\Vec{R}_{Null}$. Hyper-parameters specific to \modelname\ are the number of stages $K$, the number of slots $M$, and the
   number of rules $N$. $W^q$, $\hat{W}^k$, $W^{R_a}, $ $W^k$, $\widetilde{W}^k$, and $\widetilde{W}^q$ are learnable weights.

    \BlankLine 
    \item[]
    \For{each input element $\vx^t$ with $t \gets 1$ to $T$}  {
    
    \vspace{2mm}
    \STATE{{\bfseries \itshape Step 1:} \itshape Update or infer the entity state in each slot $j$, $\vV_j^{t,0}$, from
    the previous state, $\vV_j^{t-1,K}$ and the current input $\vx_t$.}
   %\STATE{\bfseries \itshape Step 1: Representation of  different entities  $\{ \vV_1^{t}, \ldots, \ldots, \vV_M^{t}\}$ at a particular time-step $t$. Function of the input at time step $t$ as well as representation of entities at time-step $t-1$ i.e., $\{ \vV_1^{t-1}, \ldots, \ldots, \vV_M^{t-1}\}$ } 
    %\vspace{2mm}
   %\STATE{\bfseries \itshape Repeat Procedure consisting of slot-rule pair selection to identify primary slot (Step 2), seeking information from other slots i.e., selecting contextual slot (Step 3), and applying the selected rule to update the state of primary slot  (Step 4) for K stages. } \\

    \item[]
   \STATE{{\bfseries \itshape Step 2:} \itshape Select the set of primary slots $\vP$}
   \STATE \bull $\vR_a = Concatenate([\Vec{R_i} \quad \forall \vi \in \{1, \ldots, N\}]) \vW^{R_a}$
   \STATE \bull $\vk_j = \vV_j^{t} \hat{\vW}^k \quad \forall \vj \in \{1, \dots, M\}$
   \STATE \bull $\vP = \{\vj \quad \text{if} \quad \vR_{a}\vk_j + \gamma > \vR_{Null}\vk_j + \gamma, \quad \text{where} \quad j \in \{1, \ldots M\}\quad \text{and}\quad \gamma \sim \mathrm{Gumbel} (0,1)  \}$
   %\vW^q
   
    \item[]
   \STATE{{\bfseries \itshape Step 3:} \itshape Select a rule for each primary slot in $\vP$}
  \STATE \bull $\vk_{i} = \Vec{\vR}_{i} \vW^k  \quad \forall i \in \{1, \ldots, N\}$
  \STATE \bull $\vq_p = \vV_{p}^{t} \vW^q \quad \forall p \in \vP$
  \STATE \bull $\vr_p = \mathrm{argmax}_i (\vq_p\vk_i + \gamma) \quad \forall \vi \in \{1, \ldots, N\} \quad \forall \vp \in \vP, \quad \text{where} \quad \gamma \sim \mathrm{Gumbel} (0,1)$
  
  \item[]
   \STATE{{\bfseries \itshape Step 4:} \itshape Select a  contextual slot for each primary slot}
   
   \STATE \bull $\vq_p = \vV_p^t\widetilde{\vW}^q \quad \forall \vp \in \vP$
   \STATE \bull $\vk_j = \vV_{j}^{t} \widetilde{\vW}^k \quad \forall j \in \{1, \ldots, M\}$
   \STATE \bull $c_p = \mathrm{argmax}_j (\vq_{p} \vk_j + \gamma) \quad \forall \vj \in \{1, \ldots, M\} \quad \forall \vp \in \vP,\quad \gamma \sim \mathrm{Gumbel} (0,1)  $ 
   
   \item[]
   
   \STATE{{\bfseries \itshape Step 5:} \itshape Apply selected rule to each primary slot conditioned on the contextual slot}
   
   \STATE \bull $\widetilde{\vR}_p = \mathrm{MLP}_{\vr_p}(\mathrm{Concatenate}([\vV_{p}^{t}, \vV_{c_p}^{t}])) \quad \forall \vp \in \vP$
   
   \STATE \bull $\vV_{p}^{t+1} = \vV_{p}^{t} + \widetilde{\vR}_p \quad \forall \vp \in \vP$

   }
   
   %\item[]
   
   %\STATE{\bfseries \itshape Step 4: Apply the selected rule $\vR_{\hat{\vi}}$ to the selected entity $\vV_{\hat{\vj}}^{t}$ by taking into account relevant entity $\vV_{\hat{\vr}}^{t}$ }
   
   %\STATE \bull $\widetilde{\vR} = \mathrm{MLP}_{\hat{\vi}}(\mathrm{Concatenate}([\vV_{\hat{\vj}}^{t}, \vV_{\hat{\vr}}^{t}]))$
   
   %\STATE \bull $\vV_{\hat{\vj}}^{t} = \vV_{\hat{\vj}}^{t} + \widetilde{\vR}$
   %\EndFor
   
   %\item[]
   %\STATE{\bfseries \itshape Step 5: Repeat Procedure consisting of entity-rule pair selection (Step 2), seeking information from other relevant entity  (Step 3), and applying the selected rule  (Step 4).}

\end{algorithmic}
\end{algorithm}

\section{Related Work}

\begin{table*}
\small
\centering
\caption{Learning representation of entities, dynamics of those entities and interaction between entities: Different ways in which the previous work has learned representation of different entities as a set of slots, dynamics of those entities (i.e., whether different entities follow the same dynamics, different dynamics or in a dynamic way i.e context dependent manner) and how these entities interact with each other. We note that the proposed model is agnostic as to how one learn the representation of different entities, as well as how these entities behave (i.e., dynamics of those entities). }
\begin{tabular}{|l|l|l|l|}
\hline
Relevant Work & Entity Encoder  & Entity Dynamics &  Entity Interactions \\ \hline
RIMs \citep{goyal2019recurrent} & Interactive Enc. & Different & Dynamic \\
\hline
%R-NEM \citep{van2018relational} & Top-Down Encoder & Graph Neural Network (+Attention) \\ 
%\hline
MONET \citep{burgess2019monet} & Sequential Enc. & NA & NA \\
\hline 
IODINE \citep{greff2019multi} & Iterative Reconstructive Enc. & NA  & NA \\
\hline
C-SWM \citep{kipf2019contrastive} & Bottom-Up Enc. &  NA & GNN \\ 
\hline
OP3 \citep{veerapaneni2020entity} & Iterative Reconstructive Enc. & Same  & GNN \\
\hline
SA \citep{locatello2020object} & Iterative Interactive Enc. & NA  & NA \\ 
\hline
SCOFF \citep{goyal2020object} & Interactive Enc. & Dynamic &  Dynamic \\
\hline
\end{tabular}
\end{table*}

%\paragraph{Production Systems in connectionist research:} Production systems were one of the first attempts to model cognitive behaviour and form the basis of many existing models of cognition. However, in traditional symbolic AI, both the key entities and the rules that operated on the entities were given. However, for AI agents such as robots trying to make sense of their environment, the only observables are low-level variables like pixels in images. To generalize well, an agent must induce high-level entities directly form the pixel level data as well as discover the rules as to how these entities interact with each other \cite{Bengio-consciousness-arxiv2017,  goyal2020inductive, ke2021systematic}.

%Fedrick and Mozer (1994) have proposed a neural net categorization model  whose feature extraction stage remaps a high dimensional input feature space  into a lower dimensional "psychological" space, and then classifies the input  based on the psychological representation.  The feature extractor is a standard feedforward neural net, which feeds into ALCOVE \citep{kruschke1992alcove}, a  categorization model.  They show that computationally, this sort of architecture  can produce better learning and generalization performance than ALCOVE alone  or a multilayered feedforward neural network alone.

\cite{mcmillan1991connectionist} have studied a neural net model, called RuleNet, that learns simple string-to-string mapping rules.  RuleNet consists of two components:  a feature extractor and a set of simple  condition-action rules -- implemented in a neural net -- that operate on the extracted features.  Based on a training set of input-output examples, RuleNet performs better than a standard neural net architecture in which the processing is completely unconstrained.
%To partition images into entities, structural inductive biases, e.g., slot-based mechanisms have been proposed.  Graph Neural Networks have been prominently used to capture interactions between these entities \citep{scarselli2008graph, bronstein2017geometric, watters2017visual, raposo2017discovering,santoro2017simple, gilmer2017neural, van2018relational, kipf2018neural, battaglia2018relational, tacchetti2018relational}. 

%Physical processes in the world often have a modular structure which human cognition appears to exploit, with complexity emerging through combinations of simpler subsystems. Machine learning seeks to uncover and use regularities in the physical world. Although these regularities manifest themselves as statistical dependencies, they are ultimately due to dynamic processes governed by causal physical phenomena. These processes are mostly evolving independently and only interact sparsely

\textbf{Key-Value Attention.} Key-value attention \citep{bahdanau2014neural} defines the backbone of updates to the slots in the proposed model. This form of attention is widely used in Transformer models \citep{vaswani2017attention}. Key-value attention selects an input value based on the match of a query vector to a key vector associated with each value. To allow easier learnability, selection is soft and computes a convex combination of all the values. Rather than only computing the attention once, the multi-head dot product attention mechanism (MHDPA) runs through the scaled dot-product attention multiple times in \emph{ parallel}. There is an important difference with  \modelname: in MHDPA, one can treat different heads  as different rule applications. Each head (or rule) considers \emph{all} the other entities as relevant arguments as compared to the sparse selection of arguments in \modelname.

%\textbf{Modularity and Neural Networks}.  A network can be composed of several modules, each meant to perform a distinct function, and hence can be seen as a combination of experts \citep{jacobs1991adaptive, bottou1991framework,ronco1996modular, reed2015neural, andreas2016neural,rosenbaum2017routing, fernando2017pathnet, shazeer2017outrageously, kirsch2018modular, rosenbaum2019routing, lamb2020neural} routing information through a gated activation of modules. The framework  can be stated as having a meta-controller $c$ which from a particular state $s$, selects a particular expert or rule $a = c(s)$ as to how to transform the state $s$. These works generally assume that only a single expert (i.e., winner take all) is active at a particular time step. Such approaches factorize knowledge as a set of experts (i.e a particular expert is chosen by the controller). Whereas in the proposed work, there's a factorization of knowledge both in terms of entities as well as rules (i.e., experts) which act on these entities.

%This is the primary motivation behind \modelname.  
\textbf{Sparse and Dense Interactions.} GNNs model  pairwise interactions between all the slots hence they can be seen as capturing \textit{dense} interactions \citep{scarselli2008graph, bronstein2017geometric, watters2017visual,  van2018relational, kipf2018neural, battaglia2018relational, tacchetti2018relational}. Instead, verbalizable interactions in the real world are sparse~\citep{bengio2017consciousness}: the immediate effect of an action is only on a small subset of entities. In \modelname, a selected rule only updates the state of a subset of the slots hence the interactions in \modelname are sparse.  

\textbf{Modularity and Neural Networks}.  A network can be composed of several modules, each meant to perform a distinct function, and hence can be seen as a combination of experts \citep{jacobs1991adaptive, bottou1991framework,ronco1996modular, reed2015neural, andreas2016neural,rosenbaum2017routing, fernando2017pathnet, shazeer2017outrageously, kirsch2018modular, rosenbaum2019routing, lamb2020neural} routing information through a gated activation of modules. The framework  can be stated as having a meta-controller $c$ which from a particular state $s$, selects a particular expert or rule $a = c(s)$ as to how to transform the state $s$. These works generally assume that only a single expert (i.e., winner take all) is active at a particular time step. Such approaches factorize knowledge as a set of experts (i.e a particular expert is chosen by the controller). Whereas in the proposed work, there's a factorization of knowledge both in terms of entities as well as rules (i.e., experts) which act on these entities.

\textbf{Graph Neural Networks.} GNNs model  pairwise interactions between all the entities hence they can be termed as capturing \textit{dense} interactions \citep{scarselli2008graph, bronstein2017geometric, watters2017visual,  van2018relational, kipf2018neural, battaglia2018relational, tacchetti2018relational}. Interactions in the real world are sparse. Any action affects only a subset of entities as compared to all entities. For instance, consider a set of bouncing balls, in this case a collision between 2 balls $a$ and $b$ only affects $a$ and $b$ while other balls follow their default  dynamics. Therefore, in this case it may be useful to model only the interaction between the 2 balls that collided (sparse) rather than modelling the interactions between all the balls (dense). This is the primary motivation behind \modelname.

In the \modelname, one can view the resultant computational graph as a result of sequential application of rules as a GNN, where the states of the entities represent the different nodes, and different rules dynamically instantiate an edge between a set of entities, again chosen in a dynamic way. Its important to emphasize that the topology of the graph induced in the \modelname\ is dynamic, while in most GNNs the topology is fixed. Through thorough set of experiments, we show that learning sparse and dynamic interactions using \modelname\ indeed works better than learning dense interactions using GNNs. We show that NPS outperforms state-of-the-art GNN-based architectures such as C-SWM \cite{kipf2019contrastive} and OP3 \cite{op3veerapaneni} while learning world-models.

\textbf{Neural Programme Induction.} Neural networks have been studied as a way to address the problems of learning procedural behavior and program induction \citep{graves2014neural, reed2015neural, neelakantan2015neural, cai2017making, xu2018neural, trask2018neural, bunel2018leveraging, li2020strong}. The neural network parameterizes a policy distribution $p(a|s)$, which induces such a controller, which issues an instruction $a = f(s)$ which has some pre-determined semantics over how it transforms $s$. Such approaches also factorize knowledge as a set of experts. Whereas in the proposed work, there's a factorization of knowledge both in terms of entities as well as rules (i.e., experts) which act on these entities. \cite{evans2019making} impose a  bias in the form of rules which is used to to define the state transition function, but we believe both the rules and the representations of the entities can be learned from the data with sufficiently strong inductive bias.

\textbf{RIMs, SCOFF and \modelname}. \cite{goyal2019recurrent, goyal2020object} are a key inspiration for our work. RIMs consist of ensemble of modules sparingly interacting with each other via a bottleneck of attention. Each RIM module is specialized for a particular computation and hence different modules operate according to different dynamics. RIM modules are thus not interchangeable. \cite{goyal2020object} builds upon the framework of RIMs to make the slots interchangeable, and allowing different slots to follow similar dynamics. In SCOFF,  the interaction different entities is via direct entity to entity interactions via attention, whereas in the \modelname\ the interactions between the entities are mediated by sparse rules i.e., rules which have consequences for only a subset of the entities.

\section{NPS Specific Parameters}
We use query and key size of 32 for the attention mechanism used in the selection process in steps 2 and 3 in algorithm \ref{alg:neural_production_systems}. We use a gumbel temperature of 1.0 whenever using gumbel softmax. Unless otherwise specified, the architecture of the rule specific MLP is shown in table \ref{tab:rule_mlp}.

\begin{table}[]
 \renewcommand{\arraystretch}{1.5}
    \centering
    \begin{tabular}{|c|c|c|}
    \hline
         Type & Size & Activation  \\
         \hline
         Linear & 128 & ReLU \\ 
         Linear & Slot Dim. i.e size of $\vV$ & \\
    \hline
    \end{tabular}
    \caption{Architecture of the rule-specific MLP ($MLP_{r}$) in algorithm \ref{alg:neural_production_systems}.}
    \label{tab:rule_mlp}
\end{table}

\section{MNIST Transformation Task} \label{mnist_appendix}
\begin{table}
\centering
    \begin{tabular}{|c|c|c|c|c|}
    \hline
    &Type & Channel & Activation & Stride \\
    \hline
    \multirow{4}{*}{Encoder} & Conv2D $[4 \times 4]$ & 16 & ELU & 2 \\
    &Conv2D $[4 \times 4]$ & 32 & ELU & 2 \\
    &Conv2D $[4 \times 4]$ & 64 & ELU & 2 \\
    &Linear & 100 & ELU & - \\
    \hline
    \multirow{7}{*}{Decoder} & Linear & 4096 & ReLU & - \\
    & Interpolate (scale factor = 2)& - & - & -  \\
    & Conv2D $[4 \times 4]$ & 32 & ReLU & 1 \\
    & Interpolate scale factor = 2 & - & - & - \\
    & Conv2D $[4 \times 4]$ & 16 & ReLU & 1 \\
    & Interpolate scale factor = 2 & - & - & - \\
    & Conv2D $[3 \times 3]$ & 1 & ReLU & 1 \\
    \hline
    \end{tabular}
    \caption{The architecture of the convolutional encoder and decoder for the MNIST Transformation task.}
    \label{tab:mnist_enc_dec}.

\end{table}

The two entities are first encoded in two slots $M=2$: the first entity (the image) is encoded with a convolutional encoder to a $d$-dimensional vector; the second entity (the one-hot operation vector) is mapped to a $d$-dimensional vector using a learned weight matrix. Note that this step is different than Step 1 of Algorithm \ref{alg:neural_production_systems} since we don't have multiple timesteps here and we use the transformation embedding as one of the slots which is not done for any of the other experiments. We feed these two slots to the \modelname\ module. Step 2 in algorithm \ref{alg:neural_production_systems} will match the transformation embedding to the corresponding rule. Step 3 in algorithm \ref{alg:neural_production_systems} can be used to select the correct slot for rule application (i.e. the image ($\vX$) representation). Step 4 can then apply the MLP of the selected rule to the selected slot from step 3 and output the result, which is passed through a common decoder to generate the transformed image. We train the model using binary cross entropy loss.

As highlighted before, NPS has 3 components: (1) Rule selection, (2) Entity selection, and (3) Dynamic edge instantiation. The main motivation behind using mnist transformation task is to study the rule selection aspect of NPS. Therefore, we have only 1 entity and study whether NPS can learn 4 different rules to represent the 4 operations and learn to used them correctly. We observe that NPS is indeed able to do so. 

\begin{figure}[t]
    \centering
    \includegraphics[width = 8cm]{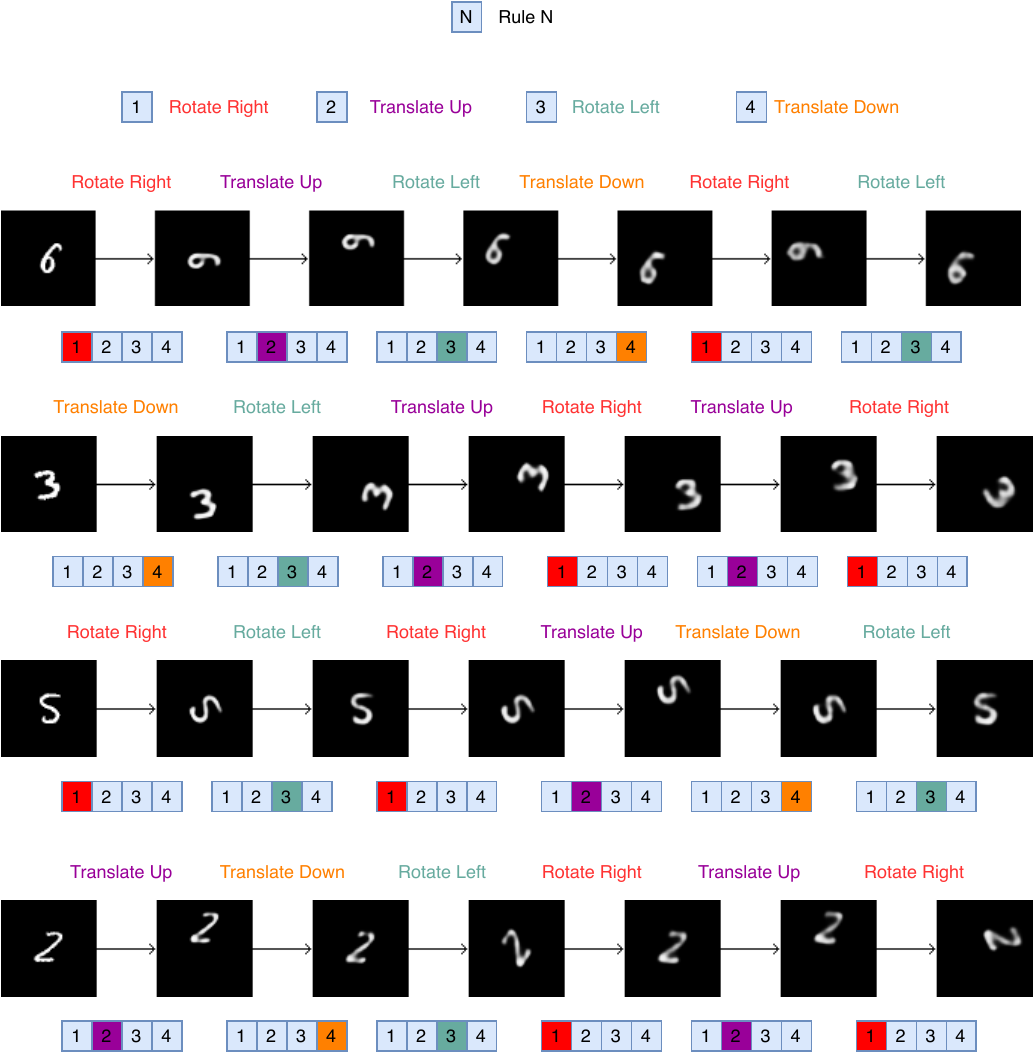}
    \caption{\textbf{MNIST Transformation Task}. Demonstration of \modelname\ on the MNIST transformations task. The proposed model can be used to compose any novel combination of transformations on any digit by hand-picking the rule vector that was assigned to each corresponding operation during training.} % and applying it to the encoded representation of the digit $\vZ$.}
    \label{fig:synthetic_mnist}
\end{figure}

In this task, we use images of size $64 \times 64$. There are 4 possible transformations that can be applied on an image: [Translate Up, Translate Down, Rotate Left, and Rotate Right]. During training, we present an image and the corresponding operation vector and train the model to output the transformed image. We use binary cross entropy loss as our training objective. As mentioned before, we observe that NPS learns to assign a separate rule to each transformation. We can use the learned model to perform and compose novel transformations on MNIST digits. For example, if we want to perform the \textit{Rotate Right} operation on a particular digit, we input the one-hot vector specifying the \textit{Rotate Right} operation alongwith the digit to the learned model. The model than outputs the resultant digit after performing the operation. A demonstration of this process is presented in Figure \ref{fig:synthetic_mnist}. Figure \ref{fig:synthetic_mnist} shows the actual outputs from the model. 

\textbf{Setup}. We first encode the image using the convolutional encoder presented in table \ref{tab:mnist_enc_dec}. We present the one-hot transformation vector and the encoded image representation as slots to NPS (algorithm \ref{alg:neural_production_systems}). NPS applies the rule MLP corresponding to the given transformation to the encoded representation of the image which is then decoded  to give the transformed image using the decoder in table \ref{tab:mnist_enc_dec}. We use a batch size of 50 for training. We run the model for 100 epochs. This experiment takes 2 hours on a single v100.

\section{Coordinate Arithmetic Task} \label{appendix:coordinate_arithmetic}

This task is mainly designed to test whether NPS can learn all the operations in the environment correctly when the operation to be performed is not provided as input and whether it can select the correct entity to apply the operation to. The task is structured as follows: We first sample a pair random coordinates $X = [ (x_i, y_i), (x_j, y_j) ]$. The expected output $Y = [(\hat{x_i}, \hat{y_i}), (\hat{x_j}, \hat{y_j})]$ is obtained by performing a randomly selected operation on a randomly selected coordinate (hereafter referred to as "primary coordinate") from the input. Therefore, one of the coordinates from the expected output $Y$ has the same value as the corresponding coordinate in the input $X$ while the other coordinate (primary coordinate) will have a different value. Since each defined operation takes 2 arguments, we also need to select another coordinate (hereafter referred to as "contextual coordinate"), to perform the operation on the primary coordinate. This contextual coordinate is selected randomly from the input $X$. For example, if the index of the primary coordinate is $j$, the index of the contextual coordinate is $i$, and the selected transformation is \textit{Y Subtraction}, then the expected output $Y = [(x_i, y_i), (x_j, y_j - y_i)])$.

%We provide the input $X$ as well as the expected output $Y$ to \modelname. Using this information, \modelname\ first picks the primary slot (primary coordinate) and the rule to be used using step 2 of algorithm \ref{alg:neural_production_systems}. Then, \modelname\ picks the contextual slot (contextual coordinate) using step 3 of algorithm \ref{alg:neural_production_systems}. Then, \modelname\ applies the rule MLP of the selected rule to the primary slot conditioned on the contextual slot as in step 4 of algorithm \ref{alg:neural_production_systems} to get the final output. Note that the expected output is used only for selection of \{primary slot, contextual slot, rule\} triplet. The rule MLP (which produces the final output) does not have access to the expected output hence the model cannot copy over the expected output $Y$, provided as the input, to produce the final output. 

We use an NPS model with 4 rules. Each coordinate is a 2-dimensional vector. The slots that are input to algorithm \ref{alg:neural_production_systems} consist of a set of 4 coordinates (2 input coordinates and their corresponding output coordinates): $\vX = [ (x_i, y_i), (x_j, y_j) ]$ and $Y = [(\hat{x_i}, \hat{y_i}), (\hat{x_j}, \hat{y_j})]$. We concatenate the input coordinates with the corresponding output coordinates to form 4-dimensional vectors: $\vX = [(x_i, y_i, \hat{x_i}, \hat{y_i}), (x_j, y_j, \hat{x_j}, \hat{y_j})]$. These make up the 2 slots that are input algorithm \ref{alg:neural_production_systems}. Note that the slots are hand designed and not obtained using any convolution encoder. This is the main difference between this implementation and algorithm \ref{alg:neural_production_systems}. The remaining steps of algorithm \ref{alg:neural_production_systems} are followed as is. These 2 slots are used in step 2 and step 3 of algorithm \ref{alg:neural_production_systems} to select the \{primary slot, rule\} pair and the contextual slot. While applying the rule MLP (i.e. step 4 of algorithm \ref{alg:neural_production_systems}), we only use the input coordinates and discard the output coordinates: $\vX = [ (x_i, y_i), (x_j, y_j) ]$. We use a rule embedding dimension of 12. We use 16 as the intermediate dimension of the Rule MLP. We also use dropout with $p = 0.35$ on the selection scores in subpart 3 of step 2 in algorithm \ref{alg:neural_production_systems}.

For the baseline, we use similar rule MLPs as in \modelname\ and replace the {primary slot, contextual slot, rule} selection procedure by a routing MLP similar to \cite{swicth_transformers}. The routing MLP consists of a 4-layered MLP with intermediate dimension of 32 interleaved with Relu activations. The input to this MLP is an 8-dimension vector consisting of both the slots mentioned in the previous paragraph. The output of this MLP is passed to 3 different linear layers: one for selecting the primary slot, one for selecting the contextual slots, and one for selecting the rule. We then apply the rule MLP of the selected rules to the corresponding slots. Here again, while applying the rule MLP we discard the outputs and only use the inputs.  

We evaluate the model on 2 criterion: (1) Whether it can correctly recover all available operations from the data and learn to use a separate rule to represent each operation. (2) The mean-square error between the actual output and expected output.

\textbf{Setup}. We generate a training dataset of 10000 examples and a test dataset of 2000 examples. We train the model for 300 epochs using a batch size of 64. We use adam optimizer for training with a learning rate of 0.0001. Training takes 15 minutes of a single GPU. 
% 5 rules
%0 {0: 360, 1: 99, 2: 45, 3: 13, 4: 0}
%1 {0: 0, 1: 57, 2: 15, 3: 99, 4: 335}
%2 {0: 0, 1: 482, 2: 0, 3: 1, 4: 0}
%3 {0: 0, 1: 39, 2: 453, 3: 2, 4: 0}

% 6 rules
%0 {0: 65, 1: 34, 2: 0, 3: 322, 4: 0, 5: 96}
%1 {0: 41, 1: 66, 2: 296, 3: 0, 4: 103, 5: 0}
%2 {0: 478, 1: 5, 2: 0, 3: 0, 4: 0, 5: 0}
%3 {0: 8, 1: 486, 2: 0, 3: 0, 4: 0, 5: 0}

\section{Parallel vs Sequential Rule Application}
\subsection{Shapes Stack}

\begin{figure}
    \centering
    \includegraphics[width = 14cm]{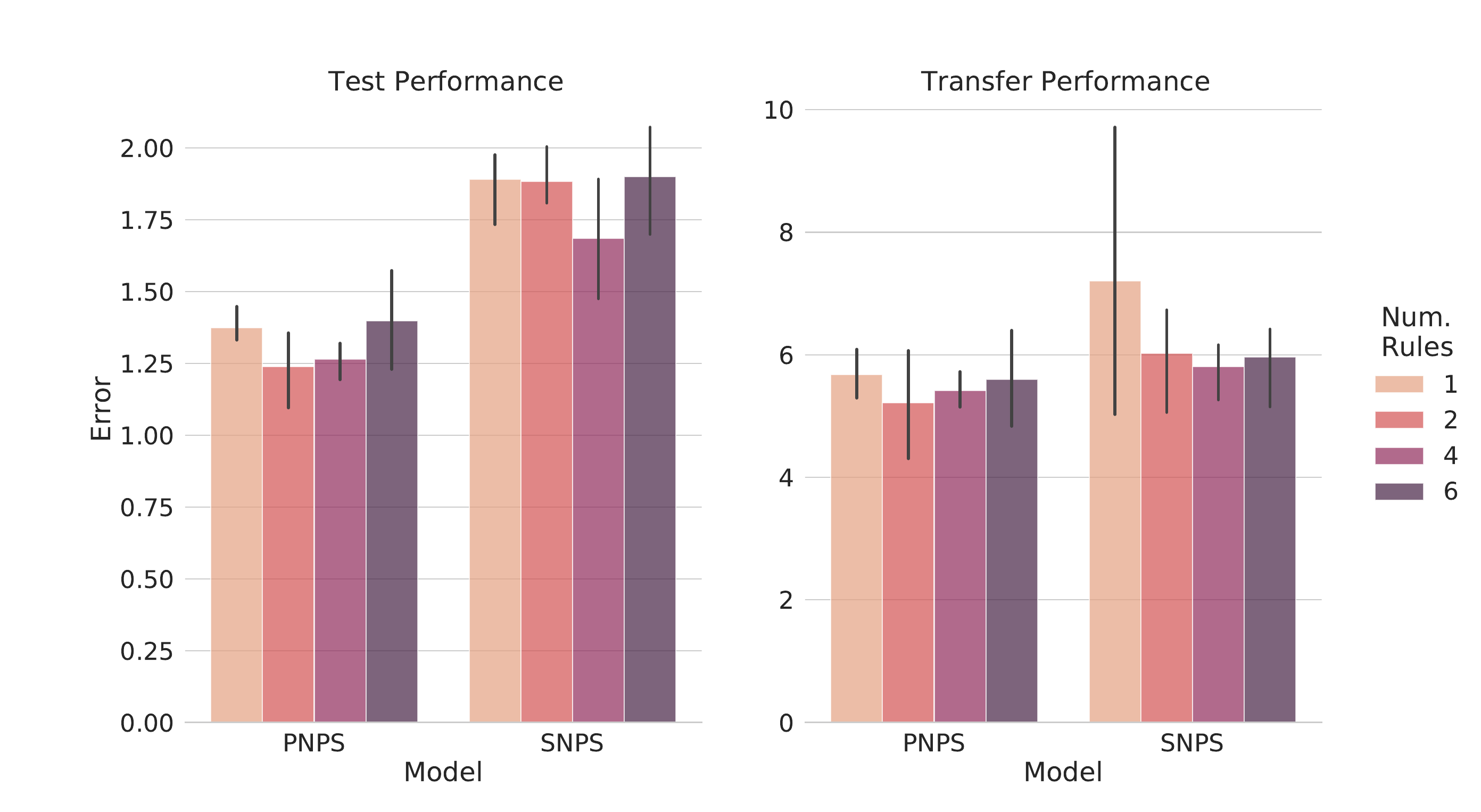}
    \caption{Here we analyse the effect of Number of Rules  on both PNPS and SNPS in the shapes stack environment. We observe that there is an optimal number of rules in both cases which is 2 for PNPS and 4 for SNPS.}
    \label{fig:ss_num_rules}
\end{figure}

\textbf{Effect of Number of Rules}. We also study the effect of number of rules on both PNPS and SNPS in the test and transfer settings. Figure \ref{fig:ss_num_rules} shows the results of our analysis. We can see that there is an optimal number of rules $\vN = \vk$ for both PNPS and SNPS and the performance drops for $\vN < \vk$ and $\vN > \vk$. We can see that  $\vk = 2$ for PNPS and $\vk = 4$ for SNPS. The drop in performance for $\vN < \vk$ can be attributed to lack of capacity (i.e. number of rules being less than the required number of rules for the environment). Consequently, the drop in performance for $\vN > \vk$ can be attributed to the availability of more than the required number of rules. The extra rules may serve as noise during the training process. 

\textbf{Training Details}. We train the model for 1000000 iterations using a batch size of 20. The training takes 24 hours for PNPS and 48 hours for SNPS. We use a single v100 gpu for each run. We set the rule embedding dimension to 64.

\begin{wraptable}{r}{5.5cm}
\vspace{-2mm}
\centering
\scriptsize	
\centering 
\renewcommand{\arraystretch}{1.3}
\setlength{\tabcolsep}{1.5pt}
\begin{tabular}{|c|c|c|}
\hline
        Model Name & Test &   Transfer   \\
        \hline
        OP3 & \g{0.32}{0.04} & \g{0.14}{0.1}  \\
        \hline
        SNPS  & \highlight{\g{0.51}{0.07}} & \highlight{\g{0.34}{0.08}} \\
        \hline
\end{tabular}
\caption{This table shows the comparison of the proposed SNPS models against the OP3 baseline in terms of ARI scores (higher is better) on the bouncing balls task. SNPS replaces the GNN used in OP3 by rules.}  
\label{tab:op3_comparison}
\vspace{-1.1\baselineskip}
\end{wraptable}

\subsection{Bouncing Balls}\label{appendix:bouncing_balls}

\begin{figure}
    \centering
    \includegraphics[width = 14cm]{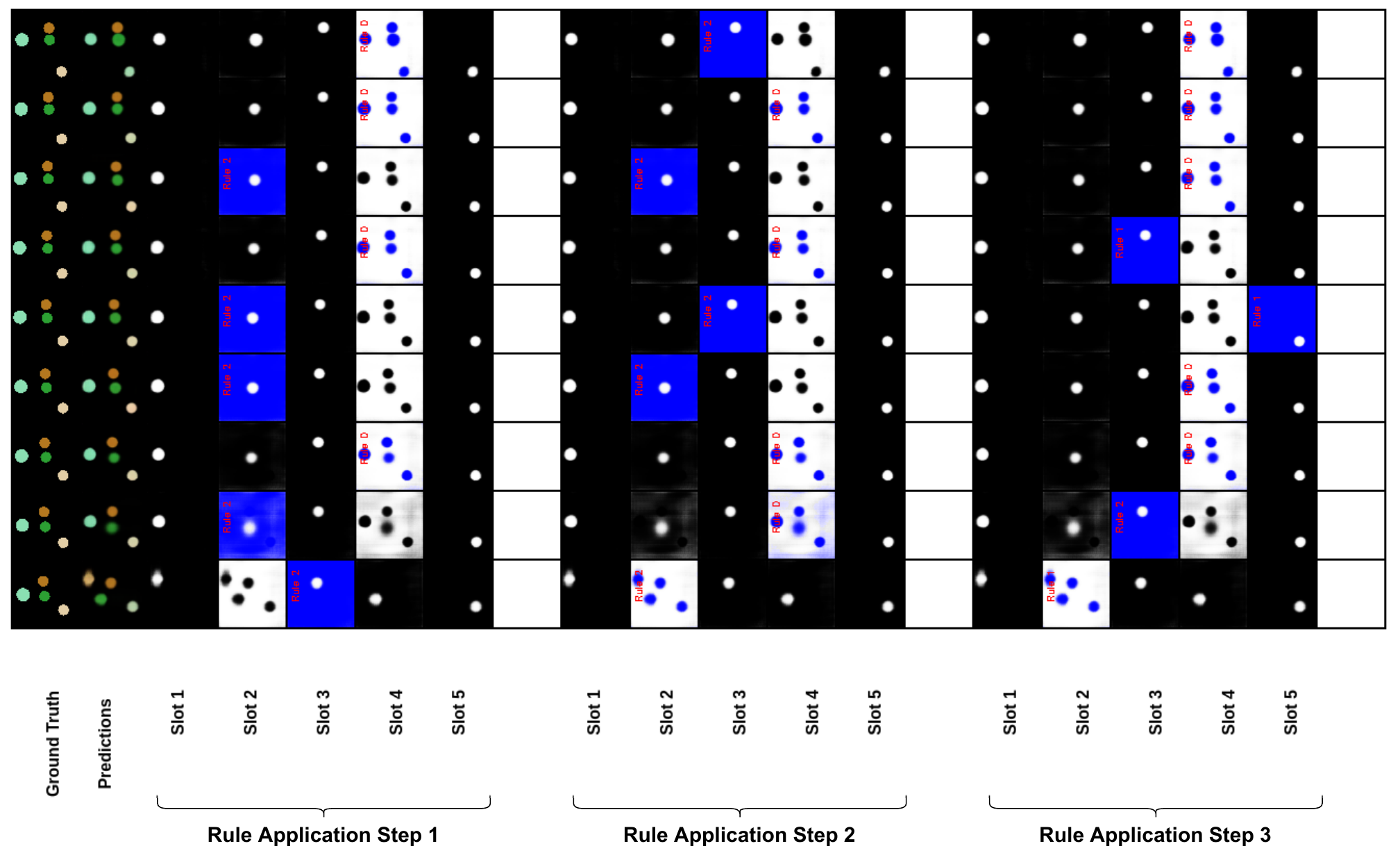}
    \caption{ In the figure, we use an NPS model with 3 rules and 3 rule application steps. We analyze the entity and rule selection by NPS per time step. A rule application on a slot is shown by highlighting that slot with blue color. The index of the applied rule is also mentioned in the slot. We can see that whenever a rule is applied on the slot representing the background, rule 0 is used. On the other hand, whenever it is applied on the slots representing one of the balls, rule 1 or rule 2 are used. We can also see that rules are mainly only being applied to the two balls in the middle that are touching or close to touching while no rules are being applied to the ball on the top since it stays constant throughout this episode. The ball at the bottom is also mostly constant and receives only 1 rule application when its close to colliding with the wall.}
    \label{fig:bb_vis}
\end{figure}

\textbf{Setup}. We consider a bouncing-balls environment in which multiple balls move with billiard-ball dynamics. We validate our model on a colored version of this dataset.   This task is setup as a next step prediction task where the model is supposed to predict the motion of a set of balls that follow billiard ball dynamics.  The output of each of our models consists of a seperate binary mask for each object in a frame along with an rgb image corresponding to each mask which contains the rgb values for the pixels specified by that mask. During training each of the balls can have one of the four possible colors, and during testing we increase the number of balls from 4 to 6-8.

We use the following baselines for this task:
\begin{itemize}
    \item \textbf{SCOFF \cite{goyal2020object}}: This factorizes knowledge in terms of object files that represent entities and schemas that represent dynamical knowledge. The object files compete to represent the input using a top-down input attention mechanism. Then, each object file updates its state using a particular schema which it selects using an attention mechanism. %For the proposed model, we use the object file states as the slots that are input into NPS. The NPS algorithm is applied after every object file update step.
    \item \textbf{SCOFF++}: Here we use SCOFF with 1 schema. We replace the input attention mechanism in SCOFF with an iterative attention mechanism as proposed in slot attention \cite{locatello2020objectcentric}. We note that slot attention proposes to use iterative attention by building on  the idea of top-down attention as proposed in \citep{goyal2019recurrent}. We note that slot attention was only evaluated on the static images. Here, the query is a function of the hidden state of the different object files in SCOFF from the previous timestep, which allows temporal consistency in slots across the video sequence. 
    \vspace{-0.2cm}
    %\item \textbf{OP3}: We use the OP3 model from \cite{op3veerapaneni}. We follow the exact same setup as \cite{weis2020unmasking}.
    
\end{itemize}

For our instantiation of NPS, we replace pairwise communication attention in SCOFF++ with PNPS or SNPS. From the discussion in Section \ref{section:exp_par_seq} we find that SNPS outperforms the SCOFF-based model and PNPS.

We further test the proposed SNPS model against another strong object-centric baseline called OP3 (\cite{op3veerapaneni}). We follow the exact same setup as \cite{weis2020unmasking}. For the proposed model we replace the GNN in OP3 with SNPS. We present the results of this comparison in Table \ref{tab:op3_comparison}. We can see that SNPS comfortably outperforms OP3. 

An intuitive visualization of the rule and entity selection for the bouncing balls environment is presented in Figure \ref{fig:bb_vis}.

\textbf{Training Details}. We run each model for 20 epochs with batch size 8 which amounts to 24 hours on a single v100 gpu. For the OP3 experiment, we use 5 rules and 3 rule application steps.  We use a rule embedding dimension of 32 for our experiments.

\subsection{Discussion of parallel vs sequential NPS} \label{parvsseq}
We have introduced two forms of NPS - Parallel and Sequential. Parallel NPS offers lower application sparsity as compared Sequential NPS while both have the same contextual sparsity. The same contextual sparsity indicates that the interactions or rules learned by both SNPS and PNPS are of same capacity since the rules in both take 2 arguments (primary slot and contextual slot). SNPS has the favourable property of being able to compose multiple rules by virtue of multiple rule application steps. We find that PNPS works better in environments where the nature of interactions are inherently dense and frequent which is expected due to the lower application sparsity of PNPS while SNPS works better in environments where interactions are much more rare. One caveat of this approach is that we need to know the structure of the environment beforehand to make an informed decision on whether to use SNPS or PNPS. Ideally, we would want an algorithm that can learn to use PNPS or SNPS depending on the environment. We leave this exploration for future work.

\section{Benefits of Sparse Interactions Offered by NPS}

\subsection{Sprites-MOT: Learning Rules for Physical Reasoning}\label{appendix:spmot_metrics}

 \textbf{Setup}. We use the OP3 model \cite{op3veerapaneni} as our baseline for this task. We follow the exact same setup as \cite{weis2020unmasking}. To test the proposed model, we replace the GNN-based transition model in OP3 with the proposed \modelname. The output of the model consists of a seperate binary mask for each object in a frame along with an rgb image corresponding to each mask which contains the rgb values for the pixels specified by the mask.

\textbf{Evaluation protocol}. We use the same evaluation protocol as followed by \cite{weis2020unmasking} which is based on the MOT (Multi-object tracking) challenge \cite{DBLP:journals/corr/MilanL0RS16}.  To compute these metrics, we have to match the objects in the predicted masks with the objects in the ground truth mask. We consider a match if the intersection over union (IoU) between the predicted object mask and ground truth object mask is greater than 0.5. The results on these metrics can be found in Table \ref{tab:spmot}. We consider the following metrics:
\begin{itemize}
    \item \textbf{Matches (Higher is better)}: This indicates the fraction of predicted object masks that are mapped to the ground truth object masks (i.e. IoU > 0.5).
    %\vspace{-0.2cm}
    \item \textbf{Misses (Lower is better)}: This indicates the fraction of ground truth object masks that are not mapped to any predicted object masks.
    %\vspace{-0.2cm}
    \item \textbf{False Positives (Lower is better)}: This indicates the fraction of predicted object masks that are not mapped to any ground truth masks.
    %\vspace{-0.2cm}
    \item \textbf{Id Switches (Lower is better)}: This metric is designed to penalize \textit{switches}. When a predicted mask starts modelling a different object than the one it was previously modelling, it is termed a an \textit{id switch}. This metric indicates the fraction of objects that undergo and id switch.
    %\vspace{-0.2cm}
    \item \textbf{Mostly Tracked (Higher is better)}: This is the ratio of ground truth objects that have not undergone and id switch and have been tracked for at least 80\% of their lifespan.
    %\vspace{-0.2cm}
    \item \textbf{Mostly Detected (Higher is Better)}: This the ratio of ground truth objects that have been correctly tracked for at least 80\% of their lifespan without penalizing id switches.
    %\vspace{-0.2cm}
    \item \textbf{MOT Accuracy (MOTA) (Higher is better)}: This measures the fraction of all failure cases i.e. false positives, misses, and id switches as compared to the number of objects present in all frames. Concretely, MOTA is indicated by the following formula:
    \begin{equation}
        MOTA = 1 - \frac{\sum_{t = 1}^{T} M_t + FP_t + IDS_t}{\sum_{t=1}^{T}O_t}
    \end{equation}
    where, $M_t$, $FP_t$, and $IDS_t$ indicates the misses, false positives, id switches at timestep $t$ and $O_t$ indicates the number of objects at timestep $t$. 
    \item \textbf{MOT Precision (MOTP) (Higher is better)}: This metric measures the accuracy between the predicted object mask and the ground truth object mask relative to the total number of matches. Here, accuracy is measured in IoU between the predicted masks and ground truth mask. Concretely, MOTP is indicated using the following formula:
    \begin{equation}
        MOTP = \frac{\sum_{t = 1}^{T} \sum_{i = 1}^{I} d_{t}^{i}}{\sum_{t = 1}^{T}c_t}
    \end{equation}
    where, $d_{t}^{i}$ measures the accuracy for the $i^{th}$ matched object between the predicted and the ground truth mask measured in IoU. $c_t$ indicates the number of matches in timestep $t$. 
\end{itemize}
\begin{figure}
    \centering
    \includegraphics[width = 12cm]{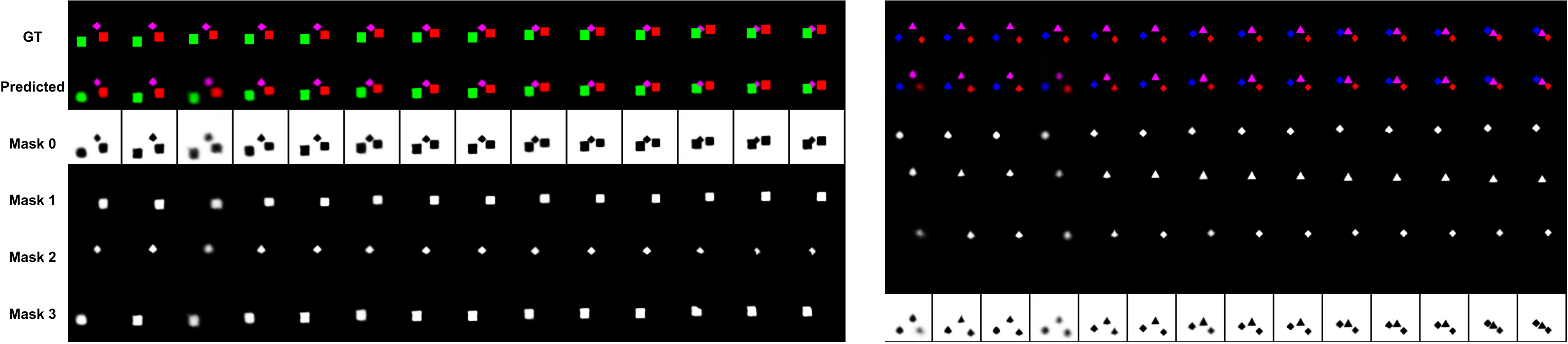}
    \caption{This figure shows the predictions of the OP3 model using the proposed NPS as a transition model. We can see that the proposed model succeeds in segregating each entity into separate slots and predicting the motion of each individual entity.}
    \label{fig:spmot}
\end{figure}

\begin{table*}[h]
\scriptsize
    \centering
    \renewcommand{\arraystretch}{1.7}
    \setlength{\tabcolsep}{1.5pt}

    \begin{tabular}{|c|c|c|c|c|c|c|c|c|c|c|}
    \hline
    Model  & MOTA $\uparrow$ & MOTP $\uparrow$ & Mostly Detected $\uparrow$ & Mostly Tracked $\uparrow$ & Match $\uparrow$ & Miss $\downarrow$ & ID Switches $\downarrow$ & False Positives $\downarrow$ \\
    \hline
    OP3 &  \g{89.1}{5.1} & \g{78.4}{2.4} & \g{92.4}{4.0} & \g{91.8}{3.8} & \g{95.9}{2.2} & \g{3.7}{2.2}& \highlight{\g{0.4}{0.0}} & \g{6.8}{2.9} \\
NPS &  \highlight{\g{90.72}{5.15}} & \highlight{\g{79.91}{0.1}} & \highlight{\g{94.66}{0.29}} & \highlight{\g{93.18}{0.84}} & \highlight{\g{96.93}{0.16}} & \highlight{\g{2.48}{0.07}} & \g{0.58}{0.02} & \highlight{\g{6.2}{3.5}} \\
%Ours & 10 & 5 & \g{90.96}{0.0} & \g{79.68}{0.01} & \g{94.06}{0.17} & \g{92.87}{0.32} & \g{96.38}{0.02} & \g{3.14}{0.02} & \g{0.48}{0.0} & \g{5.42}{0.04} \\
%Ours & 15 & 3 & 87.65 & 79.16 & 92.99 & 91.19 & 95.73 & 3.33 & 0.95 & 8.08 \\ 
\hline
    \end{tabular}
    \caption{\textbf{Sprites-MOT}. Comparison between the proposed \modelname\ and the baseline OP3 for various MOT (multi-object tracking) metrics on the sprites-MOT dataset ($\uparrow$: higher is better, $\downarrow$: lower is better). Average over 3 random seeds.}
    \label{tab:spmot}
    \vspace{-3mm}
\end{table*}

\textbf{Model output}. We show the predictions of the proposed model in figure \ref{fig:spmot}. We use 10 rules and 3 rule application steps for our experiments. We use a rule embedding dimension of 64 for our experiments. Each rule is parameterized by a neural network as described in Tab. \ref{tab:rule_mlp}.

\subsection{Physics Environment}
\begin{wrapfigure}{r}{0.5\textwidth}
   \vspace{-15mm}
   \begin{center}
     \includegraphics[scale=0.26, trim={24.5cm 0cm 0cm 0cm},clip ]{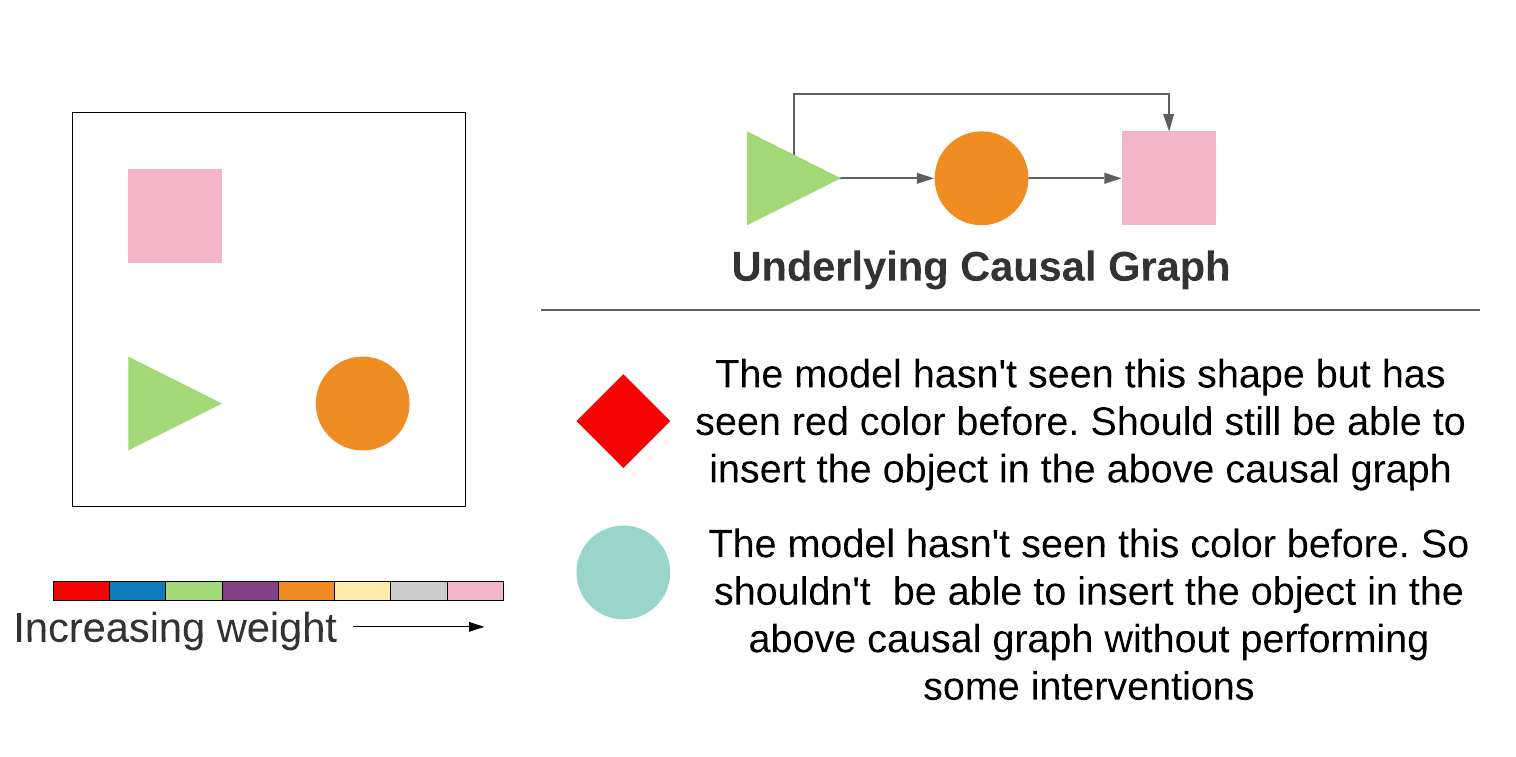}
  \end{center}
  \caption{Demonstration of the physics environment.}
  \label{fig:weighted_blocks}
\end{wrapfigure}

A demonstration of this environment can be found in Figure \ref{fig:weighted_blocks}. Each color in the environment is associated with a unique weight. The model does not have access to this information. For the model to accurately predict the outcome of an action, it needs to infer the weights from demonstrations. Inferring the correct weights will allow the model to construct the correct causal graph for each example as shown in Figure \ref{fig:weighted_blocks} which will allow it predict the correct outcome of each action in the environment. We can see that as long as the model has the correct mapping from the colors to the weights, it will be able to deal with an object of any shape irrespective of whether it has seen the shape before or not as long as it has observed the color before. Therefore, to perform well in this environment the model must infer the correct mapping from colors to weights . Learning any form of spurious correlation between the shape and the weight will penalize its performance. 

The agent performs stochastic interventions (actions) in the environment to infer the weights of the blocks. Each intervention makes a block move in any of the 4 available directions (left, right, up, and down). When an intervened block $A$ with weight $W_A$ comes into contact with another block $B$ with weight $W_B$, the block $B$ may get pushed if $W_B < W_A$ else $B$ will remain still. Hence, any interaction in this environment involves only 2 blocks (i.e., 2 entities) and the other blocks are not affected. Therefore, modelling the interactions between all blocks for every intervention, as is generally done using GNNs, may be wasteful. NPS is particularly well suited for this task as  any rule application takes into account only a subset of entities, hence considering interactions between those blocks only. Note that the interactions in this environment are not symmetrical and NPS can handle such relations. For example, consider a set of 3 blocks: $\{A_0, A_1, A_2\}$. If an intervention leads to $A_1$ pushing $A_0$, then NPS would apply the rule to the slot representing entity $A_0$. Since the movement of $A_0$ would depend on whether $A_0$ is heavier or lighter than $A_1$, NPS would also select the slot representing entity $A_1$ as a contextual slot and take it into account while applying the rule to $A_0$. Therefore, NPS can represent sparse and directed rules, which as we show, is more useful in this environment then learning dense and undirected relationships (or commutative operations).

\textbf{Data collection}. For the data, the agent performs random interventions or actions in the environment and collects the corresponding episodes. We collect 1000 episodes of length 100 for training and 10000 episodes of length 10 for evaluation.  

\textbf{Metrics}. For this task, we evaluate the predictions of the models in the latent space. We use the following metrics described in \citep{kipf2019contrastive} for evaluation: \textbf{Hits at Rank 1 (H@1)}: This score is 1 for a particular example if the predicted state representation is nearest to the encoded true observation and 0 otherwise. Thus, it measures whether the rank of the predicted representation is equal to 1 or not, where ranking is done over all reference state representations by distance to the true state representation. We report the average of this score over the test set. Note that, higher H@1 indicates better model performance. \textbf{Mean Reciprocal Rank (MRR)}: This is defined as the average inverse rank, i.e, MRR $= \frac{1}{N} \sum_{n=1}^N \frac{1}{\text{rank}_n}$ where $\text{rank}_n$ is the rank of the $n^{th}$ sample of the test set where ranking is done over all reference state representations. Here also, higher MRR indicates better performance. 

\textbf{Setup}. Here, we follow the experimental setup in \cite{ke2021systematic}. We use images of size $50 \times 50$. We first encode the current frame $\vx^t$ using a convolutional encoder and pass this encoded representation using top-down attention as proposed in \citep{goyal2019recurrent, goyal2020object} to extract the separate entities in the frame as slots. We use a 4-layered convolutional encoder which preserves the spatial dimensions of the image and encodes each pixel into 64 channels. We pass this encoded representation to the object encoder which extracts the entities in the frame as $M$ 64-sized slots ($\vV_{1 \ldots M}^t$). We then concatenate each slot with the action ($\va^t$) taken in the current frame and pass this representation to NPS which selects a rule to apply to one of the slots using algorithm \ref{alg:neural_production_systems}. The following equations describe our model in detail. We use $M = 5$ since there are 5 objects in each frame. We use a rule embedding dimension of 64 for our experiments.  

\begin{center}
\begin{tabular}{l}
    \bull $\hat{\vx^t} = \text{Encoder}(\vx^t)$  \\
    \bull $\vV_{1 ... M}^t = \text{Slot Attention}(\hat{\vx}^t)$ \\
    \bull $\vV_{i}^t = \text{Concatenate} (\vV_{i}^t, \va^t)$    $ \forall \text{i} \in \{1, \ldots, M\}$ \\
    \bull $\vV_{1 \ldots M}^{t+1} = \text{NPS} (\vV_{1 \ldots M}^{t})$
\end{tabular}
    
\end{center}

Here, NPS acts as a transition model. For the baseline, we use GNN as the transition model similar to \cite{kipf2019contrastive}.

\textbf{Training Details}. The objective of the model is to make accurate predictions in the latent space. Mathematically, given the current frame $\vx^t$ and a set of actions $\va^t, \va^{t+1}, \va^{t+2}, \ldots, \va^k$, the model performs these actions in the latent space as described in the above equations and predicts the latent state after applying these actions, i.e., $\vV_{1 \ldots M}^{t+k}$. Both the metrics, H@1 and MRR measure the closeness between the predicted latent state and $\vV_{1 \ldots M}^{t+k}$ and the ground truth latent state $\bar{\vV}_{1 \ldots M}^{t+k}$  which is obtained by passing frame $x^{t + k}$ through the convolutional encoder and slot attention module. During training we use the contrastive loss which optimizes the distance between the predicted and the ground truth representations. We train the model for 100 epochs (1 hour on a single v100 gpu). Mathematically, the training objective can be formulated as follows:

\begin{align*}
    \textit{Contrastive Training }: \argmin_{\text{Encoder, Transition}} &H + \max(0, \gamma - \tilde{H}) \\
    H &= \text{MSE}(\hat{\vV}_{1 \ldots M}^{t+1}, \vV_{1 \ldots M}^{t+1}) \\
    \tilde{H} &= \text{MSE}(\tilde{\vV}_{1 \ldots M}^{t+1}, \vV_{1 \ldots M}^{t+1}) \\
    \tilde{\vV}^{t+1} &: \text{ Negative latent state obtained from random shuffling of batch} \\
    \hat{\vV}^{t+1} &: \text{ Ground truth latent state for frame $x^{t+1}$}
\end{align*}

\begin{figure}
    \centering
    \subfigure[\textbf{Performance on MRR}]{\includegraphics[width = 7cm]{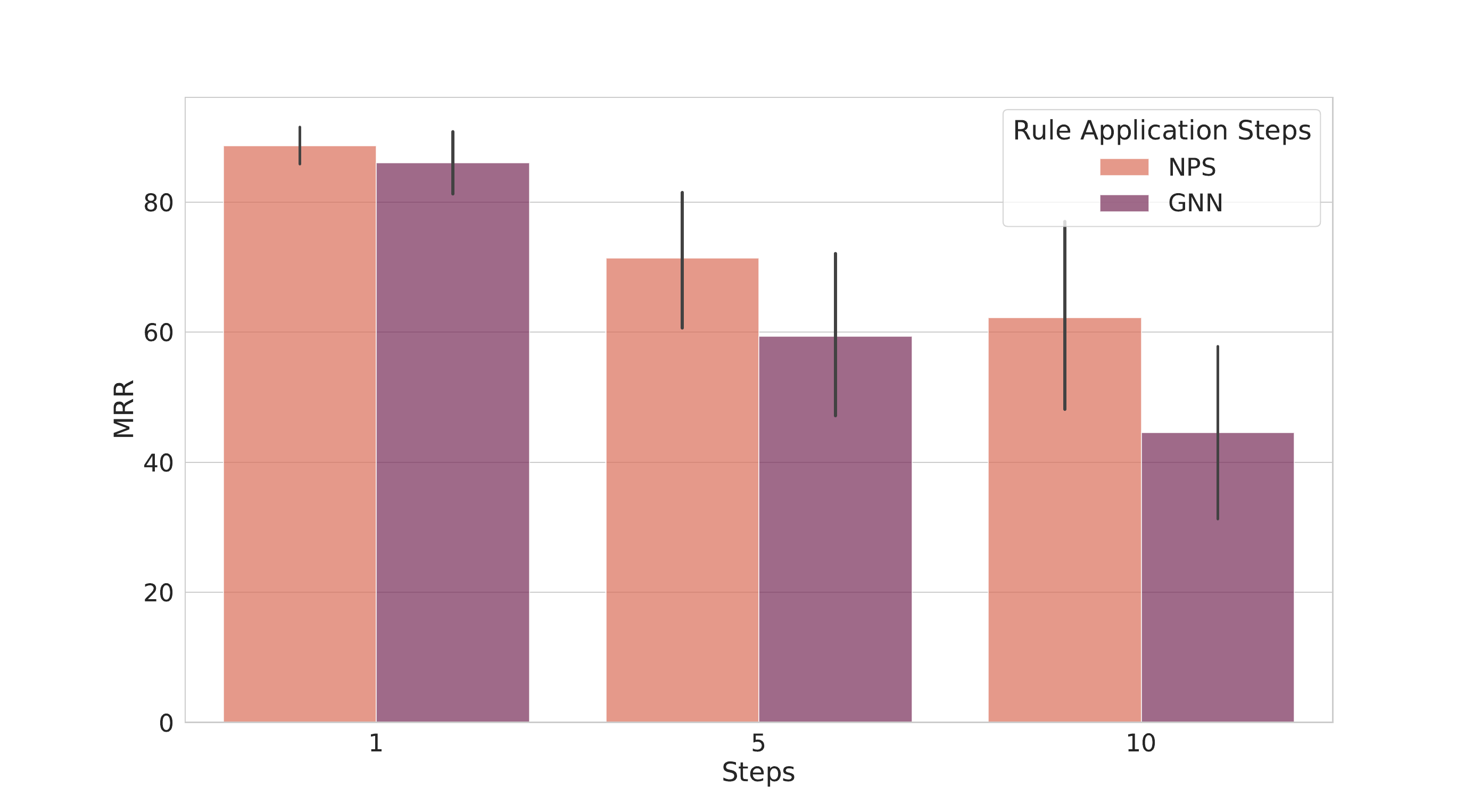}}
    \subfigure[\textbf{Effect of Rule Application Steps on H@1 and MRR}]{
    \includegraphics[width = 14cm]{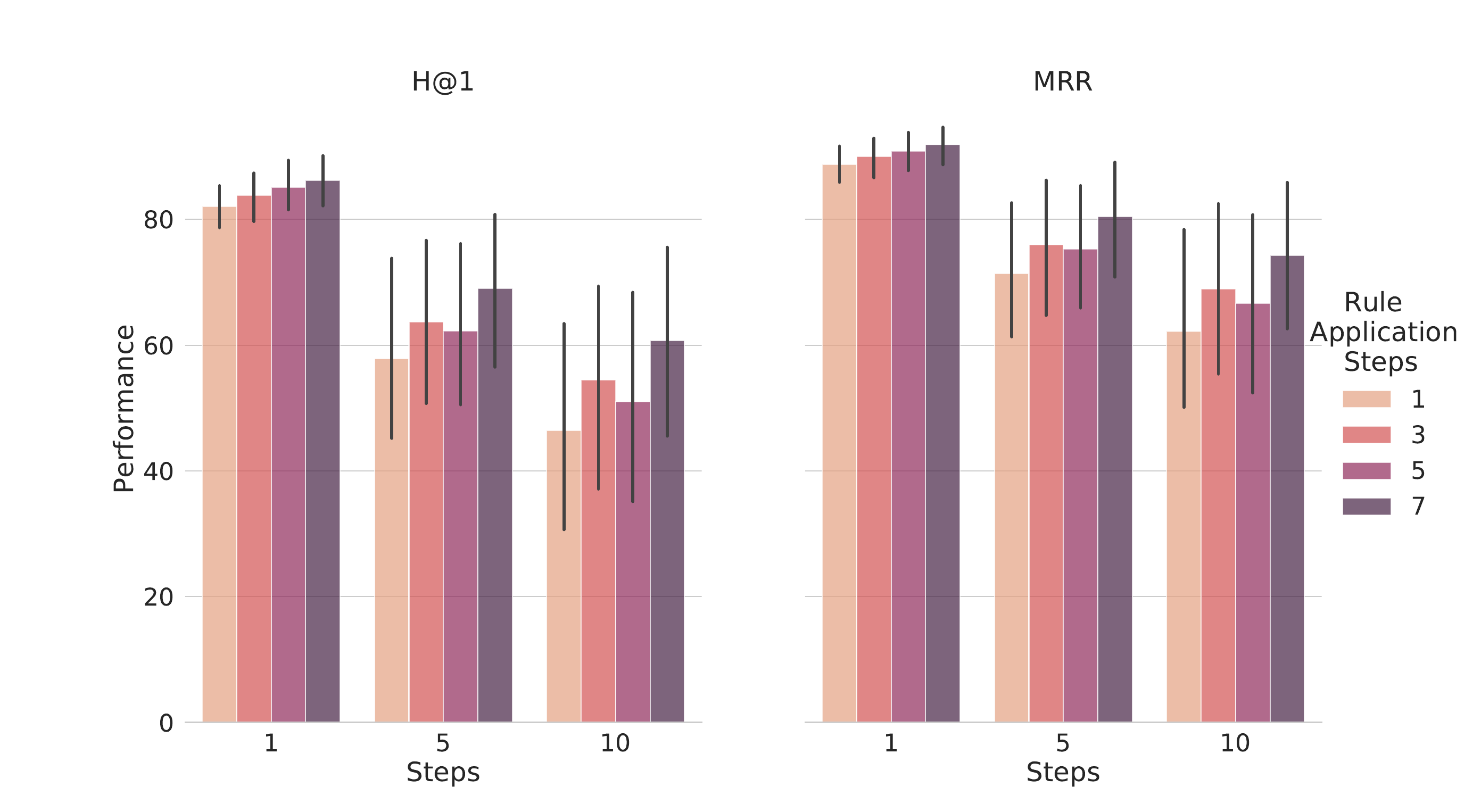}}
    \caption{\textbf{Physics Environment}. (a) Here we compare the performance of NPS and GNN on the MRR metric for various forward-prediction steps.(We use 1 rule and 1 rule application step) (b) Here we analyse the effect of the rule application steps on the H@1 and MRR metric for NPS. We can see that, in general, the performance increases as we increase the number of rule application steps.}
    \label{fig:physics_mrr}
\end{figure}

\textbf{Results on MRR}. We show the results on the MRR metric for the physics environment in Figure \ref{fig:physics_mrr}(a). From the figure, we can see that NPS outperforms GNN on the MRR metric while predicting future steps.

\textbf{Effect of Rule Application Steps}. We show the effect of varying rule application steps on the physics environment in Figure \ref{fig:physics_mrr}(b). We can see that the performance increases with increasing rule application steps. Increasing the number of rule application steps seems to help since at each rule application step only sparse changes occur. We would also like to note that increasing the rule application steps comes with the cost of increased compute time due to their sequential nature.  

%From figure \ref{fig:physics_mrr}(b) , we can see that increasing rule application steps decreases performance on MRR. This decrease in performance can be attributed to increasing density of interactions captured by increasing rule application steps as highligted before in section \ref{sec:action_condition}. We can also see that the performance trend on MRR metric is the same as that on H@1 metric shown in figure \ref{fig:gnn_experiments}[a,b]. Note that we use 1 rule for these experiments since physics environment affords only one rule , i.e., heavier blocks push lighter blocks.

%For training, we follow the same setup as \cite{kipf2019contrastive}. To summarize, we use a convolutional encoder to encode the image into a set of entities, each representing a particular block. We then use a transition model for next-step prediction in the latent space. This transition model can either be a GNN or \modelname. %The details of the loss can be found in the appendix section \ref{contrastiveloss}.

\subsection{Atari}
This task is also setup as a next step prediction task in the latent space. We follow the same setup as the physics environment for this task. We use the same H@1 and MRR metric for evaluation. We test the proposed approach on 5 games: Pong, Space Invaders, Freeway, Breakout, and QBert. 

\textbf{Data collection}. Similar to the physics environment, here also the agent performs random interventions or actions in the environment and collects the corresponding episodes. We collect 1000 episodes for training and 100 episodes for evaluation.

\begin{figure}
    \centering
   \includegraphics[width=7cm]{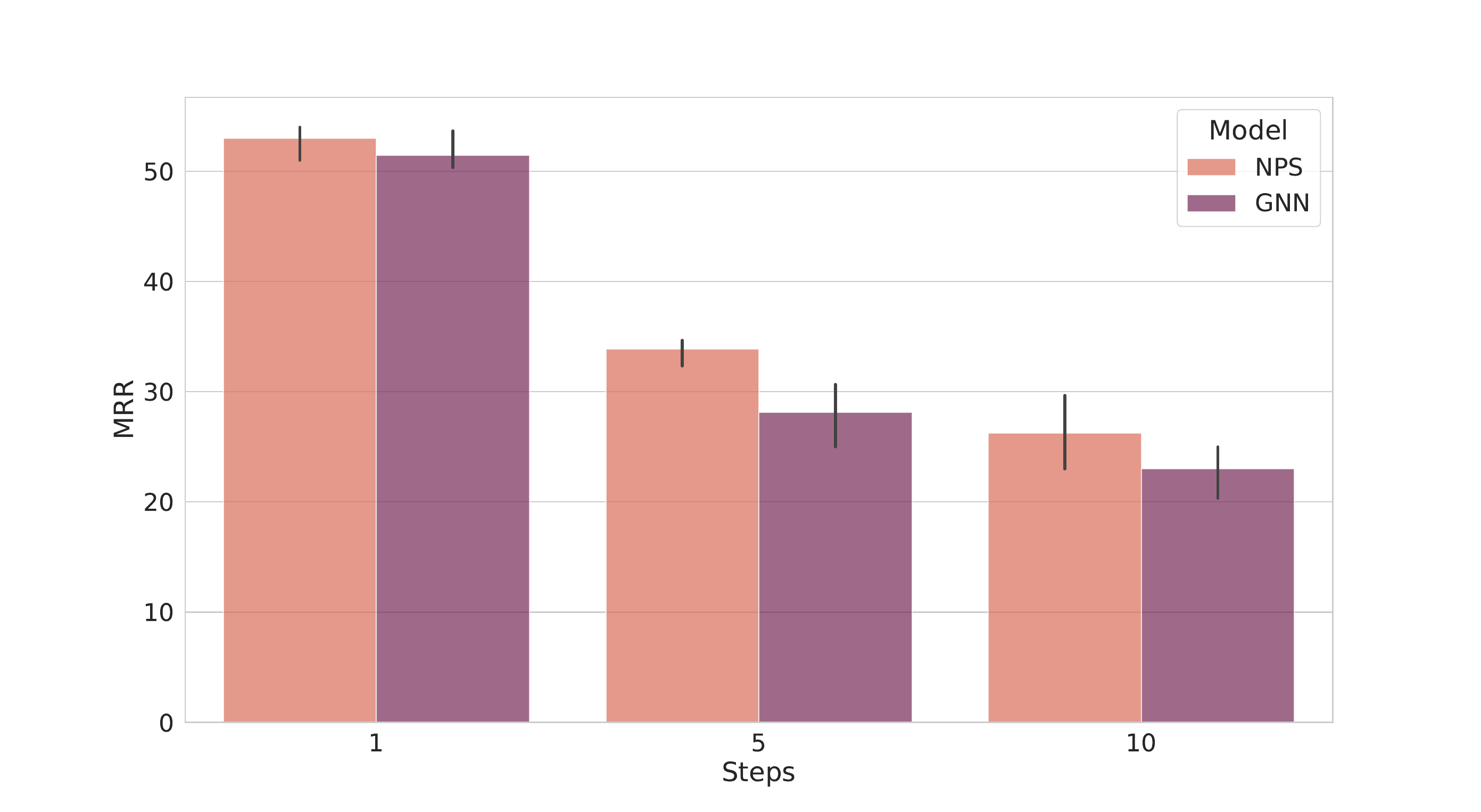}
    \caption{\textbf{Atari}. Here we compare the performance of NPS and GNN on the MRR metric for various forward-prediction steps. The results shown in the plot are averaged across 5 games.  }
    \label{fig:atari_mrr}
\end{figure}

\textbf{Performance on MRR}. We present the results on the MRR metric in figure \ref{fig:atari_mrr}. We can see that NPS outperforms on GNN for steps. We use a rule embedding dimension of 32 for our experiments.

\begin{table}
    \scriptsize
    \centering
    \renewcommand{\arraystretch}{1.7}
    \setlength{\tabcolsep}{2.2pt}

    \begin{tabular}{|l|l|l|l|l|l|l|}
    \hline
               & \multicolumn{2}{|c|}{1 Step} & \multicolumn{2}{|c|}{5 Step} & \multicolumn{2}{|c|}{10 Step} \\
        \hline
        Model, & H@1 & MRR & H@1 & MRR & H@1 & MRR  \\ \hline
        NPS & 44.09 +/- 8.30 & 60.577 +/- 7.919 & 22.25 +/- 7.567 & 38.731 +/- 9.005 & 15.615 +/- 5.748 & 30.154 +/- 7.538 \\ \hline
        GNN & 30.442 +/- 11.178 & 48.788 +/- 10.428 & 11.596 +/- 6.109 & 25.481 +/- 8.331 & 7.25 +/- 4.057 & 18.942 +/- 6.156 \\ \hline
    \end{tabular}
    \vspace{5mm}
    \caption{Here we compare the performance of NPS and GNN on the pong environment in atari. We can see that NPS convincingly outperforms GNN. Results across 50 seeds.}
\end{table}

\begin{table}
    \scriptsize
    \centering
    \renewcommand{\arraystretch}{1.7}
    \setlength{\tabcolsep}{2.2pt}

    \begin{tabular}{|l|l|l|l|l|l|l|}
    \hline
               & \multicolumn{2}{|c|}{1 Step} & \multicolumn{2}{|c|}{5 Step} & \multicolumn{2}{|c|}{10 Step} \\
        \hline
        Model, & H@1 & MRR & H@1 & MRR & H@1 & MRR  \\ \hline
        NPS & 61.269 +/- 11.576 & 74.558 +/- 9.425 & 32.827 +/- 9.179 & 52.981 +/- 9.073 & 21.385 +/- 7.118 & 39.173 +/- 8.505 \\ \hline
        GNN & 68.75 +/- 8.44 & 79.942 +/- 5.662 & 21.673 +/- 6.81 & 39.635 +/- 7.395 & 14.712 +/- 4.932 & 31.0 +/- 6.26 \\ \hline
    \end{tabular}
    \vspace{5mm}
    \caption{Here we compare the performance of NPS and GNN on the space invaders environment in atari. We can see that NPS outperforms GNN in the multiple step setting (5 and 10 step). Results across 50 seeds.}
\end{table}

%\textbf{Effect of Number of Rules}. The physics environment is a very simple environment in which all kinds of interactions occur between similarly styled blocks and there is single rule governing these interaction. Hence using 1 rule in that environment is appropriate. The atari environment is more complicated than the physics environment and interactions in atari can take place between multiple different entities. The nature of these interactions are also more complex than in the physics environment. For example, in pong when the ball hits a paddle, the resulting direction of the ball depends on the angle of interaction between the ball and paddle, the direction in which the paddle is moving, and the direction in which the ball is moving. Therefore, we hypothesize that using multiple rules will be useful in atari to capture all these complicated interactions as compared to 1 rule. We present our results in figure \ref{fig:atari_mrr}(b). We can see that NPS with 5 rules outperforms NPS with 1 rule this showing the effectiveness of multiple rules in capturing complex interactions. 

\end{document}